# Data Fusion with Latent Map Gaussian Processes

Nicholas Oune*     Jonathan Tammer Eweis-Labolle*     Ramin Bostanabad[1]

Mechanical and Aerospace Engineering,
University of California, Irvine
Irvine, California, USA

**Abstract**

Multi-fidelity modeling and calibration are data fusion tasks that ubiquitously arise in engineering design. In this paper, we introduce a novel approach based on latent-map Gaussian processes (LMGPs) that enables efficient and accurate data fusion. In our approach, we convert data fusion into a latent space learning problem where the relations among different data sources are automatically learned. This conversion endows our approach with attractive advantages such as increased accuracy, reduced costs, flexibility to jointly fuse any number of data sources, and ability to visualize correlations between data sources. This visualization allows the user to detect model form errors or determine the optimum strategy for high-fidelity emulation by fitting LMGP only to the subset of the data sources that are well-correlated. We also develop a new kernel function that enables LMGPs to not only build a probabilistic multi-fidelity surrogate but also estimate calibration parameters with high accuracy and consistency. The implementation and use of our approach are considerably simpler and less prone to numerical issues compared to existing technologies. We demonstrate the benefits of LMGP-based data fusion by comparing its performance against competing methods on a wide range of examples.

∗ Equal Contribution
[1] Corresponding author, raminb@uci.edu.

# 1    Introduction

Computer models are increasingly employed in the analysis and design of complex systems in engineering design. These models generally share at least one of the following three interconnected characteristics whose collection has motivated this work. First, they can be computationally expensive in which case they are commonly replaced/augmented with surrogates whose training can be challenged by lack of data, noise, disjoint input spaces that have both quantitative and qualitative variables, or existence of multiple responses/outputs. Second, for a particular system there are typically multiple computer models available whose fidelity is directly related to their computational costs, i.e., accurate models are generally more expensive. In such a scenario, multi-fidelity modeling techniques are used to balance cost and accuracy when using all these computer models in compute-intensive studies such as design optimization and uncertainty quantification (UQ). Third, computer models typically have some calibration parameters which are estimated by systematically comparing a model's predictions to experimental or observational data. These parameters either correspond to some properties of the underlying system being modeled or act as tuning knobs that compensate for the model deficiencies which arise from, e.g., the incorrect/simplifying assumptions built into it. In this paper, we introduce a versatile, efficient, and unified approach based on latent map Gaussian processes (LMGPs) that can be used for emulation-based multi-fidelity modeling and calibration. Henceforth, we use the term *data fusion* to refer to the collection of these two tasks because they all involve fusing or assimilating multiple sources of data.

Over the past few decades many data fusion techniques have been developed for outer-loop applications such as design optimization, sequential sampling, or inverse parameter estimation. For example, multi-fidelity modeling can be achieved via space mapping [1-3] or multi-level [4-6] techniques where the inputs of the low-fidelity data are mapped following:

$$\boldsymbol{x}_l = \boldsymbol{F}(\boldsymbol{x}_h),$$

where $\boldsymbol{x}_l$ and $\boldsymbol{x}_l$ are the inputs of low- and high-fidelity data sources, respectively, and $\boldsymbol{F}(\cdot)$ is the transformation function whose *predefined* functional form is calibrated such that $y_l\big(\boldsymbol{F}(\boldsymbol{x}_h)\big)$ approximates $y_h(\boldsymbol{x}_h)$ as closely as possible. These techniques are particularly useful in applications where higher fidelity data are obtained by successively refining the discretization of the simulation domain [4, 6], e.g., by refining the mesh when simulating the flow over an airfoil. The main disadvantage of space mapping techniques is that choosing a near-optimal functional form for $F(\cdot)$ is iterative and very cumbersome.

Two of the most important aspects of multi-fidelity modeling are choosing the emulators that surrogate the data sources and formulating the relation between these emulators. Correspondingly, several methods have been developed based on Gaussian processes (GPs) [7], Co-Kriging [8], polynomial chaos expansions [9-11], and moving least squares [12]. The interested reader is referred to [13, 14] for more comprehensive reviews on multi-fidelity modeling and how they benefit outer-loop applications.

Multi-fidelity modeling is closely related to calibration of computer models since the latter also involves working with at least two data sources where typically the low-fidelity one possesses the calibration parameters. Besides the traditional ways of estimation that are ad hoc and involve trial and error, there are more systematic methods that are based on generalized likelihood [15] or Bayesian principles [16].

Among existing methods for multi-fidelity modeling and calibration, the most popular emulator-based method in engineering design is that of Kennedy and O'Hagan (KOH) [7] which assimilates and emulates two data sources while estimating calibration parameters of the low-fidelity source (if there are any such parameters). KOH's approach is one of the first attempts that considers a broad range of uncertainty sources arising during the calibration and subsequent uses of the emulator. This approach has been used in many applications including climate simulations [17], materials modeling [18], and modeling shock hydrodynamics [19].

As we will briefly review in Section 2.2, KOH's approach assumes that the discrepancies between the two data sources are additive[2] and that both data sources as well as the discrepancy between them can be modeled via GPs. The approach then uses (fully [20, 21] or modular [18, 22-25]) Bayesian inference to find the posterior estimates of the GPs as well as the calibration parameters. The fully Bayesian version of KOH's data fusion method offers advantages such as low computational costs for small datasets or quantifying various uncertainty sources (e.g., lack of data, noise, model form error, and unknown simulation parameters). However, obtaining the joint posteriors via Markov chain Monte Carlo (MCMC) is quite effortful and expensive, especially in high-dimensions or with relatively large datasets. The modular version of KOH's approach addresses this limitation by typically using point estimates for the GP hyperparameters of the low-fidelity data [7, 23]. These estimates are obtained via maximum likelihood estimation (MLE) and, while they result in small under-estimation of uncertainties with small data, provide accurate mean predictions.

---

[2] Multiplicative terms have also been introduced to KOH's approach but are seldom adopted as they increase the identifiability issues and computational costs while negligibly improving the mean prediction accuracy.

A major limitation of KOH's approach, either fully Bayesian or modular, is that it only accommodates two data sources at a time. That is, the fusion process must be repeated $p$ times if there are $p$ low-fidelity and 1 high-fidelity data sources. In addition to being tedious and expensive, this repetitive process does not provide a straightforward diagnostic mechanism for comparing the low-fidelity sources to identify, e.g., which one(s) perform similarly or have the smallest model form error. While posterior distribution of the bias function can potentially be used for diagnostics, comparing and visualizing distributions in high dimensions is not simple.

In this paper, we aim to address the abovementioned limitations of the existing technologies for data fusion. Our primary contributions are three-fold and summarized as follows. First, we convert multi-fidelity modeling into a latent space learning problem. This conversion endows our approach with advantages such as increased accuracy, reduced costs, flexibility to jointly fuse any number of data sources, and ability to visualize correlations between data sources. This visualization allows the user to determine the optimum strategy for high-fidelity emulation by fitting LMGP only to the subset of the data sources that are well-correlated to the high-fidelity source. Second, we develop a new kernel function that enables LMGPs to not only build a probabilistic multi-fidelity surrogate but also estimate calibration parameters with high accuracy and consistency. Third, implementation of our approach is considerably simpler and less prone to numerical issues compared to existing technologies (esp. KOH's approach).

The rest of the paper is organized as follows. In Section 2, we briefly review the relevant technical background on GPs, LMGPs, and KOH's approach. In section 3, we introduce our approach to multi-fidelity modeling and calibration while demonstrating its performance on a number of analytic examples. In Section 4, we validate our approach against GPs and KOH's method on a wide range of analytic and engineering examples. We conclude the paper by discussing the advantages and limitations of our approach, considerations that should be made in its application, and its application to multi-response problems in Section 5.

## 2 Existing Technologies

We first review emulation via GPs and a variation of GPs (i.e., LMGP) for datasets that include categorical variables. Then, in Section 2.2 we summarize KOH's approach that is used in Section 4 to evaluate the performance of our data fusion approach on calibration problems. Throughout, symbols or numbers enclosed in parentheses encode sample number and are used either as subscripts or superscripts. For example, $x_{(i)}$ or $x^{(i)}$ denote the $i^{th}$ sample in a training dataset while $x_i$ indicates the $i^{th}$ component of the vector $x = [x_1, x_2, ..., x_{d_x}]^T$. In Section 2.2 we use $h$ and $l$ either as superscript or subscript to denote high- and low-fidelity data sources. For instance, $x_h^{(i)}$ and $y_h^{(i)}$ denote, respectively, the inputs and output of the $i^{th}$ sample in the high-fidelity dataset. In cases where there is more than one low-fidelity source, we add a number to the $l$ symbol, e.g., $y_{l_3}(x)$ denotes the third low-fidelity source. Lastly, we distinguish between the data source (or the underlying function) and samples by specifying the functional dependence (e.g., $y(x)$ is a function while $y$ and $\boldsymbol{y}$ are, respectively, a scalar and a vector of values).

### 2.1 Emulation via Latent Map Gaussian Processes

Denote the inputs and outputs of a system by $d_x$-dimensional vector $x = [x_1, x_2, ..., x_{d_x}]^T$ and the scalar $y$, respectively, and assume the training data come from a realization of a GP defined as:

$$\eta(x) = f(x)\beta + \xi(x),$$

where $f(x) = [f_1(x), ..., f_h(x)]$ are a set of pre-determined parametric basis functions (e.g., $x_1^2 + x_2$, $x_2^2 sin(x_1) + log(x_1 x_2), ...$), $\beta = [\beta_1, ..., \beta_h]^T$ are the unknown coefficients of the basis functions, and $\xi(x)$ is a zero-mean GP. Since $\xi(x)$ is zero-mean, it is completely characterized by its parameterized covariance function:

$$cov(\xi(x), \xi(x')) = c(x, x') = \sigma^2 r(x, x'), \qquad \text{Eq. 1}$$

where $\sigma^2$ is the process variance and $r(\cdot,\cdot)$ is a user-defined parametric correlation function. There are many types of correlation functions [26, 27], but the most common one is the Gaussian correlation function defined as:

$$r(\pmb{x},\pmb{x}') = exp\left\{-\sum_{i=1}^{d_x} 10^{\omega_i}(x_i - x'_i)^2\right\} = exp((\pmb{x}-\pmb{x}')^T \pmb{\Omega}_x (\pmb{x}-\pmb{x}')),\qquad \text{Eq. 2}$$

where $\pmb{\omega} = [\omega_1, ..., \omega_{d_x}]^T$, $-\infty < \omega_i < \infty$ are the roughness or scale parameters (in practice the ranges are limited to $-10 < \omega_i < 6$ to ensure numerical stability) and $\pmb{\Omega}_x = diag(10^{\pmb{\omega}})$. $\sigma^2$ and $\pmb{\omega}$ are collectively referred to as the hyperparameters of $r(\cdot,\cdot)$.

The correlation function in Eq. 2 depends on the distance between two arbitrary input points $\pmb{x}$ and $\pmb{x}'$. Hence, traditional GPs cannot accommodate categorical inputs (such as gender, zip code, country, material coating type, etc.) as the distance between them is not directly defined. This issue is well established in the literature, and there exist a number of strategies that address it by reformulating the covariance function such that it can handle categorical variables [28-31]. In this paper we use LMGPs [32] which are recently developed and shown to out-perform previous methods.

Let us denote the categorical inputs by $\pmb{t} = [t_1, ..., t_{d_t}]^T$ where the total number of distinct levels for qualitative variable $t_i$ is $m_i$. For instance, $t_1 = \{92697, 92093\}$ and $t_2 = \{math, physics, chemistry\}$ are two categorical inputs that encode zip code ($m_1 = 2$ levels) and course subject ($m_2 = 3$ levels), respectively. Inputs for mixed (numerical and categorical) training data are collectively denoted by $\pmb{u} = [\pmb{x}; \pmb{t}]$ which is a column vector of size $(d_x + d_t) \times 1$. To handle mixed inputs, LMGP maps categorical variables to some points in a quantitative latent space. This mapping allows to use any standard correlation function such as the Gaussian which is reformulated as follows for mixed inputs:

$$r(\pmb{u},\pmb{u}') = exp\{-\|\pmb{z}(\pmb{t}) - \pmb{z}(\pmb{t}')\|_2^2 - (\pmb{x}-\pmb{x}')^T \pmb{\Omega}_x (\pmb{x}-\pmb{x}')\},\qquad \text{Eq. 3}$$

where $\|\cdot\|_2$ denotes the Euclidean 2-norm and $\pmb{z}(\pmb{t}) = [z_1(\pmb{t}), ..., z_{d_z}(\pmb{t})]_{1\times d_z}$ is the to-be-learned latent space point corresponding to the particular combination of the categorical variables denoted by $\pmb{t}$. To find these points in the latent space, LMGP first assigns a unique vector (i.e., a prior representation) to each combination of categorical variables. Then, it uses matrix multiplication to map each of these vectors to a point in a latent space of dimensionality $d_z$:

$$z(t) = \zeta(t)A, \qquad \text{Eq 4}$$

where $\zeta(t)$ is the $1 \times \sum_{i=1}^{d_t} m_i$ unique prior vector representation of $t$ and $A$ is a $\sum_{i=1}^{d_t} m_i \times d_z$ matrix that maps $\zeta(t)$ to $z(t)$. In this paper, we use $d_z = 2$ since it simplifies visualization and has also been shown to provide sufficient flexibility for learning the latent relations [32]. We can construct $\zeta$ in a number of ways, see [32] for more information on selecting the priors. In this paper we use a form of one-hot-encoding. Specifically, we first construct the $1 \times m_i$ vector $v^i = [v_1^i, v_2^i, \ldots, v_{m_i}^i]$ for the categorical variable $t_i$ such that $v_j^i = 1$ when $t_i$ is at level $j$ and $v_j^i = 0$ when $t_i$ is at level $k \neq j$ for, $k \in 1, 2, \cdots, m_i$. Then, we set $\zeta(t) = [v^1, v^2, \cdots, v^{d_t}]$. For instance, in the above example with two categorical variables, $t_1 = \{92697, 92093\}$ and $t_2 = \{math, physics, chemistry\}$, we encode the combination $t = [92093, physics]^T$ by $\zeta(t) = [0, 1, 0, 1, 0]$ where the first two elements encode zip code while the rest encode the subject.

To emulate via LMGP, point estimates of the hyperparameters $A$, $\beta$, $\omega$, and $\sigma^2$ must be determined based on the data. These estimates can be found via either cross-validation (CV) or MLE. Alternatively, Baye's rule can be applied to find posterior distributions of the hyperparameters if prior knowledge is available. In this paper, MLE is employed because it provides a high generalization power while minimizing the computational costs [27, 33]. MLE works by estimating $A$, $\beta$, $\omega$, and $\sigma^2$ such that they maximize the likelihood of $n$ training data being generated by $\eta(x)$, that is:

$$[\widehat{\beta}, \hat{\sigma}, \widehat{\omega}, \widehat{A}] = \underset{\beta, \sigma^2, \omega, A}{\operatorname{argmax}}\ |2\pi\sigma^2 R|^{-\frac{1}{2}} \times exp\left\{\frac{-1}{2}(y - F\beta)^T(\sigma^2 R)^{-1}(y - F\beta)\right\},$$

Or equivalently,

$$[\widehat{\beta}, \hat{\sigma}, \widehat{\omega}, \widehat{A}] = \underset{\beta, \sigma^2, \omega, A}{\operatorname{argmin}}\ \frac{n}{2}log(\sigma^2) + \frac{1}{2}log(|R|) + \frac{1}{2\sigma^2}(y - F\beta)^T R^{-1}(y - F\beta), \qquad \text{Eq. 5}$$

where $R$ and $\hat{\sigma}^2$ are now functions of both $\omega$ and $A$, $log(\cdot)$ is the natural logarithm, $|\cdot|$ denotes the determinant operator, $y = [y_{(1)}, \ldots, y_{(n)}]^T$ is the $n \times 1$ vector of outputs in the training data, $R$

is the $n \times n$ correlation matrix with the $(i,j)^{th}$ element $R_{ij} = r(\pmb{x}_{(i)}, \pmb{x}_{(j)})$ for $i, j = 1, \ldots, n$, and $\pmb{F}$ is the $n \times h$ matrix with the $(k,l)^{th}$ element $F_{kl} = f_l(\pmb{x}_{(k)})$ for $k = 1, \ldots, n$ and $l = 1, \ldots, h$. By setting the partial derivatives with respect to $\pmb{\beta}$ and $\sigma^2$ to zero, their estimates can be solved in terms of $\pmb{\omega}$ and $\pmb{A}$ as follows:

$$\widehat{\pmb{\beta}} = [\pmb{F}^T \pmb{R}^{-1} \pmb{F}]^{-1} [\pmb{F}^T \pmb{R}^{-1} \pmb{y}], \qquad \text{Eq. 6}$$

$$\hat{\sigma}^2 = \frac{1}{n}(\pmb{y} - \pmb{F}\widehat{\pmb{\beta}})^T \pmb{R}^{-1}(\pmb{y} - \pmb{F}\widehat{\pmb{\beta}}), \qquad \text{Eq. 7}$$

Plugging these estimates into Eq. **5** and removing the constants yields:

$$[\widehat{\pmb{\omega}}, \widehat{\pmb{A}}] = \underset{\omega, A}{\operatorname{argmin}}\ n\log(\hat{\sigma}^2) + \log(|\pmb{R}|) = \underset{\omega, A}{\operatorname{argmin}}\ L, \qquad \text{Eq. 8}$$

By minimizing $L$ one can solve for $\widehat{\pmb{A}}$ and $\widehat{\pmb{\omega}}$, and subsequently obtain $\widehat{\pmb{\beta}}$ and $\hat{\sigma}^2$ using Eq. **6** and Eq. **7**. While many heuristic global optimization methods exist such as genetic algorithms [34] and particle swarm optimization [35], gradient-based optimization techniques based on, e.g., the L-BFGS algorithm [36], are generally preferred due to their ease of implementation and superior computational efficiency [26, 37]. With gradient-based approaches, it is essential to start the optimization via numerous initial guesses to improve the chances of achieving global optimality.

After obtaining the hyperparameters via MLE, the following closed-form formula is used to predict the response at any $\pmb{x}^*$:

$$\mathbb{E}[y^*] = \pmb{f}(\pmb{x}^*)\widehat{\pmb{\beta}} + \pmb{g}^T(\pmb{x}^*)\pmb{V}^{-1}(\pmb{y} - \pmb{F}\widehat{\pmb{\beta}}),$$

where $\mathbb{E}$ denotes expectation, $\pmb{f}(\pmb{x}^*) = [f_1(\pmb{x}^*), \ldots, f_h(\pmb{x}^*)]$, $\pmb{g}(\pmb{x}^*)$ is an $n \times 1$ vector with the $i^{th}$ element $c(\pmb{x}_{(i)}, \pmb{x}^*) = \hat{\sigma}^2 r(\pmb{x}_{(i)}, \pmb{x}^*)$, and $\pmb{V}$ is the covariance matrix with the $(i,j)^{th}$ element $\hat{\sigma}^2 r(\pmb{x}_{(i)}, \pmb{x}_{(j)})$. The posterior covariance between the responses at the two inputs $\pmb{x}^*$ and $\pmb{x}'$ is:

$$cov(y^*, y') = c(\pmb{x}^*, \pmb{x}') - \pmb{g}^T(\pmb{x}^*)\pmb{V}^{-1}\pmb{g}(\pmb{x}') + \pmb{h}(\pmb{x}^*)(\pmb{F}^T \pmb{V}^{-1} \pmb{F})^{-1} \pmb{h}(\pmb{x}')^T,$$

where $\pmb{h}(\pmb{x}^*) = (\pmb{f}(\pmb{x}^*) - \pmb{F}^T \pmb{V}^{-1} \pmb{g}(\pmb{x}^*))$.

The above formulations can be easily extended to cases where the dataset is noisy. GPs (and hence LMGPs) can address noise and smoothen data by using a nugget or jitter parameter, $\delta$, which is incorporated into the correlation matrix. That is, $R$ becomes $R_\delta = R + \delta I_{n\times n}$ where $I_{n\times n}$ is the identity matrix of size $n \times n$. If the nugget parameter is used, the estimated (stationary) noise variance in the data will be $\delta \hat{\sigma}^2$. The version of LMGP used in this paper finds only one nugget parameter and uses it for all categorical combinations, i.e., we assume that the noise level is the same for each dataset. LMGP can be modified in a straightforward manner to have a separate nugget parameter (and hence separate noise estimate) for each categorical combination.

## 2.2 KOH's Approach

KOH place certain prior distributions on the low- and high-fidelity data, calibration parameters (if there are any), and how the data sources are related. These priors are then updated using Bayes' rule to find the joint posterior distribution for the calibration parameters as well as the hyperparameters of the emulators that surrogate the low-fidelity source and the discrepancy function. In particular, they assume the following relationship between two data sources:

$$y_h(x) = y_l(x, \theta^*) + \delta(x) + \varepsilon \qquad \text{Eq. 9}$$

where $y_h(x)$ is the high-fidelity data source, $y_l(x, \theta)$ is the low-fidelity data source, $\delta(x)$ is the bias or discrepancy function, $x$ are the inputs to the system, $\theta^*$ are the unknown true calibration parameters ($\theta$ refers to the known values used as inputs in a simulation), and $\varepsilon$ denotes noise following $N(0, \lambda)$ with $\lambda$ being the unknown variance. KOH assume that low-fidelity outputs are available at $n_l = p$ distinct settings of $X_l = \{(x_l^{(1)}, \theta^{(1)}), \cdots, (x_l^{(p)}, \theta^{(p)})\}$, i.e., we have $y_l = [y_l^{(1)}, \cdots, y_l^{(p)}]^T$ where $y_l^{(i)} = y_l(x_l^{(i)}, \theta^{(i)})$. They also presume high-fidelity outputs are available at $n_h = q$ distinct settings of $X_h = \{x_h^{(1)}, \cdots, x_h^{(q)}\}$, i.e. $y_h = [y_h^{(1)}, \cdots, y_h^{(q)}]^T$. Augmenting the high-fidelity data by their unknown true calibration parameters yields $X_h(\theta^*) = \{(x_h^{(1)}, \theta^*), \cdots, (x_h^{(q)}, \theta^*)\}$.

KOH place independent GP priors on $y_l(x, \theta)$ and $\delta(x)$ with constant means $\beta_1$ and $\beta_2$ and covariance functions $c_1(\cdot,\cdot) = \sigma_1^2 r_1(\cdot,\cdot)$ and $c_2(\cdot,\cdot) = \sigma_2^2 r_2(\cdot,\cdot)$, respectively. We denote the parameters of the covariance functions (which include $\sigma^2$ and roughness parameters) by $\psi_1$ and $\boldsymbol{\psi}_2$, respectively. Assuming the full data vector $\boldsymbol{d} = [\boldsymbol{y}_l, \boldsymbol{y}_h]^T$ of size $(p+q) \times 1$ follows a multivariate normal distribution, KOH apply Bayes' rule to obtain the joint posterior distribution of $\boldsymbol{\theta}^*$, $\boldsymbol{\beta} = [\beta_1, \beta_2]^T$, $\lambda$, $\boldsymbol{\psi}_1$, and $\boldsymbol{\psi}_2$. The distribution of $\boldsymbol{d}$ reads:

$$\mathbb{E}(\boldsymbol{d} \mid \boldsymbol{\theta}^*, \boldsymbol{\beta}, \boldsymbol{\phi}) = \boldsymbol{m}_d = \boldsymbol{H}\boldsymbol{\beta} \qquad \text{Eq. 10}$$

$$cov(\boldsymbol{d} \mid \boldsymbol{\theta}^*, \boldsymbol{\beta}, \boldsymbol{\phi}) = \boldsymbol{V}_d(\boldsymbol{\theta}^*)$$
$$= \begin{pmatrix} \boldsymbol{V}_1(\boldsymbol{X}_l) & \boldsymbol{C}_1(\boldsymbol{X}_l, \boldsymbol{X}_h(\boldsymbol{\theta}^*)) \\ \left(\boldsymbol{C}_1(\boldsymbol{X}_l, \boldsymbol{X}_h(\boldsymbol{\theta}^*))\right)^T & \boldsymbol{V}_1(\boldsymbol{X}_h(\boldsymbol{\theta}^*)) + \boldsymbol{V}_2(\boldsymbol{X}_h(\boldsymbol{\theta}^*)) + \lambda \boldsymbol{I}_q \end{pmatrix} \qquad \text{Eq. 11}$$

where $\boldsymbol{H} = \begin{bmatrix} \boldsymbol{1}_{p \times 1} & \boldsymbol{0}_{p \times 1} \\ \boldsymbol{1}_{q \times 1} & \boldsymbol{1}_{q \times 1} \end{bmatrix}_{(p+q) \times 2}$, $\boldsymbol{\phi} = [\lambda, \boldsymbol{\psi}_1, \boldsymbol{\psi}_2]$, $\boldsymbol{V}_1(\boldsymbol{X}_l)_{i,j} = c_1\left(\left(\boldsymbol{x}_l^{(i)}, \boldsymbol{\theta}^{(i)}\right), \left(\boldsymbol{x}_l^{(j)}, \boldsymbol{\theta}^{(j)}\right)\right)$, $\boldsymbol{C}_1(\boldsymbol{X}_l, \boldsymbol{X}_h(\boldsymbol{\theta}^*))_{i,k} = c_1\left(\left(\boldsymbol{x}_l^{(i)}, \boldsymbol{\theta}^{(i)}\right), \left(\boldsymbol{x}_h^{(k)}, \boldsymbol{\theta}^*\right)\right)$, $\boldsymbol{V}_1(\boldsymbol{X}_h(\boldsymbol{\theta}^*))_{k,l} = c_1\left(\left(\boldsymbol{x}_h^{(k)}, \boldsymbol{\theta}^*\right), \left(\boldsymbol{x}_h^{(l)}, \boldsymbol{\theta}^*\right)\right)$, and $\boldsymbol{V}_2(\boldsymbol{X}_h(\boldsymbol{\theta}^*))_{k,l} = c_2\left(\left(\boldsymbol{x}_h^{(k)}, \boldsymbol{\theta}^*\right), \left(\boldsymbol{x}_h^{(l)}, \boldsymbol{\theta}^*\right)\right)$. After applying Bayes' rule, the joint posterior distribution is:

$$P(\boldsymbol{\theta}^*, \boldsymbol{\beta}, \boldsymbol{\phi} \mid \boldsymbol{d}) \propto P(\boldsymbol{d} \mid \boldsymbol{\theta}^*, \boldsymbol{\beta}, \boldsymbol{\phi}) P(\boldsymbol{\theta}^*, \boldsymbol{\beta}, \boldsymbol{\phi}) \qquad \text{Eq. 12}$$

where the evidence term in Bayes' rule has been dopped. By completing the square for $\boldsymbol{\beta}$ and integration, Eq. 12 can be simplified to:

$$P(\boldsymbol{\theta}^*, \boldsymbol{\phi} \mid \boldsymbol{d}) \propto |\boldsymbol{V}_d(\boldsymbol{\theta}^*)|^{-0.5} |\boldsymbol{W}(\boldsymbol{\theta}^*)|^{-0.5} \exp\left(-\frac{1}{2}\left(\boldsymbol{d} - \boldsymbol{H}\widehat{\boldsymbol{\beta}}(\boldsymbol{\theta}^*)\right)^T \boldsymbol{V}_d^{-1}(\boldsymbol{\theta}^*) \left(\boldsymbol{d} - \boldsymbol{H}\widehat{\boldsymbol{\beta}}(\boldsymbol{\theta}^*)\right)\right) P(\boldsymbol{\theta}^*) P(\boldsymbol{\phi}) \qquad \text{Eq. 13}$$

where $\widehat{\boldsymbol{\beta}}(\boldsymbol{\theta}^*) = \begin{bmatrix} \hat{\beta}_1(\boldsymbol{\theta}^*) \\ \hat{\beta}_2(\boldsymbol{\theta}^*) \end{bmatrix} = \boldsymbol{W}(\boldsymbol{\theta}^*) \boldsymbol{H}^T \boldsymbol{V}_d^{-1}(\boldsymbol{\theta}^*) \boldsymbol{d}$ and $\boldsymbol{W}(\boldsymbol{\theta}^*) = (\boldsymbol{H}^T \boldsymbol{V}_d^{-1}(\boldsymbol{\theta}^*) \boldsymbol{H})^{-1}$.

To obtain the posterior distribution of the true calibration parameters, $P(\boldsymbol{\theta}^* \mid \boldsymbol{d})$, one needs to marginalize Eq. 13 with respect to $\boldsymbol{\phi}$ using MCMC as the distribution is typically highly non-Gaussian. However, this integration is very computationally expensive so KOH and many others [22-25, 38-41] adopt a modularized approach based on MLE that obtains point estimates for predictions and parameters rather than full joint distributions. Correspondingly, we use the modularized version of KOH's approach for evaluating the performance of our approach which is also based on MLE.

# 3 Proposed Framework for Data Fusion

We first explain our approach for multi-fidelity modeling via LMGP in Section 3.1 and then extend its correlation function to enable calibration in Section 3.2. In each section, we provide pedagogical examples to facilitate the discussions and elaborate on the benefits of the learned latent space in diagnosing the results. The notation introduced in Sec. 2 is also used here.

## 3.1 Multi-fidelity Modeling via LMGP

Using LMGP for multi-fidelity modeling is quite straightforward. Consider the case where four data sources with different levels of accuracy are available and the goal is to emulate each source while (1) having limited data, especially from the most accurate source, (2) accounting for potential noises with unknown variance, and (3) avoiding *a priori* determination of how different sources are related to each other. The last condition indicates that we do *not* know $(i)$ how the accuracy of the low-fidelity models compare to each other, and $(ii)$ if low-fidelity models have inherent discrepancy which may be additive as in Eq. **9** or not. While not necessary, we assume it is known which data source provides the highest fidelity because this source typically corresponds to either observations/experiments or a very expensive computer model.

We assume $n_h$ high-fidelity samples are available whose inputs and output are denoted by $x_h$ and $y_h$, respectively. We also presume that three low-fidelity datasets with $n_{l_1}$, $n_{l_2}$, and $n_{l_3}$ samples are obtained via three different and independent simulators with *a priori unknown* fidelity levels, i.e., we do not know which simulator provides the most/least accuracy with respect to the high-fidelity source. The inputs and outputs of these simulation datasets are denoted via $x_{l_i}$, $y_{l_i}$ where $i = 1, 2, 3$. We apply no noise to the samples in this pedagogical example, but in general this may not be the case. We use the following analytic functions to generate data:

$$y_h(x) = \frac{1}{0.1x^3 + x^2 + x + 1}, \quad -2 \leq x \leq 3 \qquad \text{Eq. 14.1}$$

$$y_{l_1}(x) = \frac{1}{0.2x^3 + x^2 + x + 1}, \quad -2 \leq x \leq 3 \qquad \text{Eq. 14.2}$$

$$y_{l_2}(x) = \frac{1}{x^2 + x + 1}, \quad -2 \leq x \leq 3 \qquad \text{Eq. 14.3}$$

$$y_{l_3}(x) = \frac{1}{x^2 + 1}, \quad -2 \leq x \leq 3 \qquad \text{Eq. 14.4}$$

where the low-fidelity sources have nonlinear bias (compare the denominators) and are *not* ordered by accuracy with respect to $y_h(x)$, see **Table 1**. We calculate accuracy using relative root mean squared error (RRMSE):

$$RRMSE\left(y_{l_i}(x)\right) \approx \sqrt{\frac{(\boldsymbol{y}_{l_i} - \boldsymbol{y}_h)^T(\boldsymbol{y}_{l_i} - \boldsymbol{y}_h)}{n \times \text{var}(\boldsymbol{y}_h)}}$$

where $\boldsymbol{y}_{l_i}$ and $\boldsymbol{y}_h$ refer to vectors containing the outputs of $y_{l_i}(x)$ and $y_h(x)$ at $n = 10{,}000$ input points sampled with Sobol sequence, and $\text{var}(\boldsymbol{y}_h)$ is the variance of $\boldsymbol{y}_h$. Note that we do not use this knowledge of relative accuracy during multi-fidelity modeling via LMGP. Rather, by only using the datasets in LMGP we aim to inversely discover this relation between the fidelity levels.

**Table 1 Accuracy of data sources:** The relative root mean squared errors (RRMSEs) of the low fidelity functions $y_{l_i}(x)$. 10000 points are used in calculating the RRMSEs.

|  | $y_{l_1}(x)$ | $y_{l_2}(x)$ | $y_{l_3}(x)$ |
|---|---|---|---|
| *RRMSE* | 0.23364 | 0.14626 | 0.72549 |

To perform data fusion with LMGP, we first append the inputs with one or more categorical variables that distinguish the data sources. We can use any number of multi-level categorical variables. That is, we can either select $(i)$ a single variable with at least as many levels as there are data sources, or $(ii)$ use a few multi-level categorical variables with at least as many level combinations as there are data sources. For example, with one categorical variable we can choose $t = \{h, l_1, l_2, l_3\}$, $t = \{1, 2, 3, 4\}$, $t = \{1, a, ab, 2\}$, or $t = \{a, b, c, d, e\}$ for our pedagogical example with four data sources (in the last case level $e$ does not correspond to any of the data sources). For the remainder of this paper, we use two strategies for choosing categorical variables, see **Figure 1**. Strategy 1 uses one categorical variable with as many levels as data sources, e.g., $t = \{a, b, c, d\}$ or $t = \{1, 2, 3, 4\}$. We add the subscript $s$ to an LMGP that uses this strategy since a single categorical variable is used to encode the data sources. Strategy 2 employs multiple categorical variables where the number of variables and their levels equals the number of data sources, e.g., $t_i = \{a, b, c, d\}$ with $i = 1, 2, 3, 4$. We place the subscript $m$ to an LMGP that uses strategy 2 to indicate that multiple categorical variables are employed. As we explain below, having more levels (or level combinations if more than one $t$ is used) than data sources provides LMGP with more flexibility to learn the relation between the sources. This flexibility comes at the expense of having a larger $A$ (i.e., more hyperparameters) and higher computational costs. As we demonstrate in Section 4, the performance of LMGP is relatively robust to this modeling choice as long as there are sufficient training samples and the number of latent positions does not greatly exceed the number of hyperparameters in $A$. Regarding the latter condition, note that when LMGP must find many latent positions with a small $A$ (i.e., a very simple map), performance may suffer due to local optimality. For example, Strategy 2 with 4 data sources results in $\Pi_{i=1}^{d_t} m_i = 4^4 = 256$ latent positions (one for each possible categorical level combination where only 4 correspond to data sources) but there are only $d_z \times \Sigma_{i=1}^{d_t} m_i = 2 \times 16 = 32$ elements in $A$. These elements are supposed to map the 256 points in the latent space such that the 4 points which encode the data sources have inter-distances that reflect the underlying relation between their corresponding data sources. Without sufficient data and regularization, the learned map may provide a locally optimal solution.

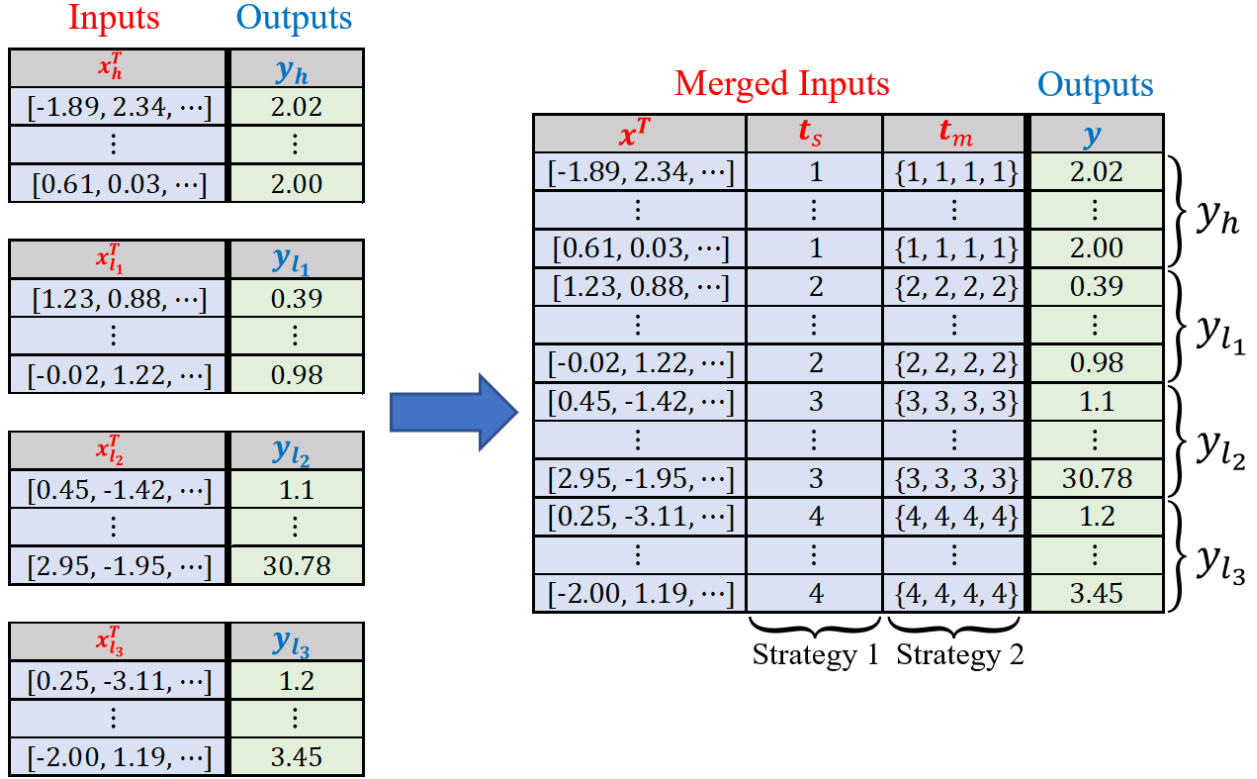

**Figure 1 Data preprocessing for multi-fidelity modeling via LMGP:** We can use any number of multi-level categorical variables when fusing data with LMGP. Shown above are two strategies for choice of $t$ for our example with four data sources. In strategy 1, we use one categorical variable with four levels (one for each data source) and assign each level to a unique data source. In strategy 2, we use a different categorical variable for each data source, and we give each categorical variable four levels (one for each data source) for a total of $4^4 = 256$ categorical combinations. We assign only four of these combinations to our data sources (only these four are enumerated in the figure), leaving 252 combinations unused. Note that while LMGP finds latent positions for these 252 combinations, the positions are not meaningful since they do not correspond to any of the data sources. The number of elements in the $A$ matrix (see Eq 4 that must be estimated for LMGP are 8 and 32 for the first and second strategies, respectively.

The above description clearly indicates that LMGP can, in principle, fuse any number of datasets *simultaneously*. In practice, this ability of LMGP is bounded by the natural limitations of GPs such as scalability to big data or very high dimensions (e.g., $dx > 50$). The recent advancements in GP modeling for big or high-dimensional data [37, 42-47] have addressed these limitations to some extent and can be directly used in LMGP for multi-fidelity modeling. However, this direction is not in the scope of this paper and will be investigated in our future works.

For the rest of this example, we select strategy 1 and append the inputs via $t = \{1, 2, 3, 4\}$ where the number of levels equals the number of data sources. We assume the datasets are highly unbalanced and use Sobol sequence to sample from the functions in Eq. 14 with $n_h = 3$ and $n_{l_1} = n_{l_2} = n_{l_3} = 20$. Upon appending, we combine the entire data into a single training dataset that is directly fed into LMGP:

$$X = \begin{bmatrix} X_h & \mathbf{1}_{n_h \times 1} \\ X_{l_1} & 2 \times \mathbf{1}_{n_{l_1} \times 1} \\ X_{l_2} & 3 \times \mathbf{1}_{n_{l_1} \times 1} \\ X_{l_3} & 4 \times \mathbf{1}_{n_{l_1} \times 1} \end{bmatrix} \text{ and } Y = \begin{bmatrix} y_h \\ y_{l_1} \\ y_{l_2} \\ y_{l_3} \end{bmatrix},$$

where $\mathbf{1}_{n \times 1}$ is an $n \times 1$ vector of ones. The fusion results are illustrated in **Figure 2 (a)** and indicate that LMGP is able to accurately emulate each data source, including $y_h(x)$ for which only three samples are provided. As illustrated in **Figure 2 (b)**, a GP fitted to only data from $y_h(x)$ provides poor performance due to lack of data.

The latent space learned by LMGP, shown in **Figure 2 (c)**, provides a powerful diagnostic tool for determining correlations between data sources without prior knowledge. As detailed in section 2.1, LMGP learns a unique latent space position for each combination of categorical variables. To understand the effect of these positions on the correlation function and hence how different data sources are related, we rewrite Eq. **3** as:

$$r(\boldsymbol{u}, \boldsymbol{u}') = \exp\{-(\boldsymbol{z} - \boldsymbol{z}')^T (\boldsymbol{z} - \boldsymbol{z}')\} \cdot \exp\{-(\boldsymbol{x} - \boldsymbol{x}')^T \boldsymbol{\Omega}_x (\boldsymbol{x} - \boldsymbol{x}')\}, \qquad \text{Eq. 15}$$

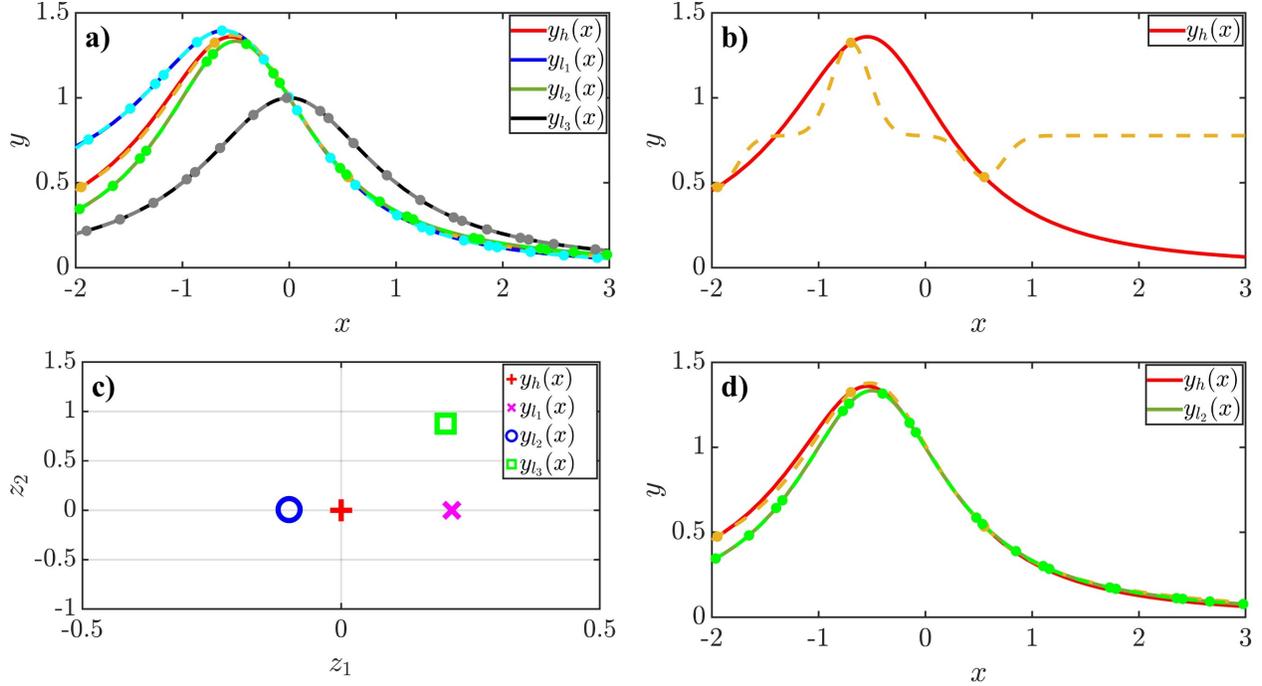

**Figure 2 Approaches to data assimilation: (a) LMGP with all available data:** LMGP fit to all available data is able to emulate each data source with high accuracy. The inaccuracy of $y_{l_3}$ does not negatively impact high-fidelity emulation performance. **(b) Standard GP:** Standard GP fit to only the three available high-fidelity samples performs poorly. **(c) Learned latent space:** LMGP only uses four datasets to learn a latent space that indicates how "close" different data sources are with respect to each other. While the datasets are quite unbalanced ($n_h = 3$ and $n_{l_1} = n_{l_2} = n_{l_3} = 20$), LMGP can clearly visualize the relative accuracy of each low-fidelity model with respect to the high-fidelity data. **(d) LMGP with only $y_{l_2}(x)$ and $y_h(x)$:** Despite the fact that $y_{l_2}(x)$ misrepresents $y_h(x)$ in some regions, LMGP is able to use correlations between the two sources to accurately emulate $y_h(x)$ with approximately equivalent accuracy to when all sources are used.

Plugging the latent positions into Eq. 15 shows that a relative distance of $\Delta \mathbf{z}^2 = (\mathbf{z} - \mathbf{z}')^T(\mathbf{z} - \mathbf{z}')$ between two points scales the correlation function by $\exp(-\Delta \mathbf{z}^2)$. Thus, we can interpret the latent space as being a distillation of the correlations between the data for each categorical combination, each of which corresponds to a different data source in multi-fidelity modeling. Note, however, that the term $\exp\{-(\mathbf{x} - \mathbf{x}')^T \mathbf{\Omega}_x (\mathbf{x} - \mathbf{x}')\}$, which accounts for the correlation between outputs at different points in the input space, remains the same as we change data sources. Thus, our modeling assumption is that this correlation is similar for all data sources. In layman's terms, we expect each data source to have a relatively similar shape. This is often true in multi-fidelity problems but is not necessarily true in the general cases (e.g., assimilating multiple responses that are uncorrelated). When our modeling assumption is not met, LMGP estimates $\mathbf{\Omega}_x$ to provide the best compromise between different sources, which may provide poor performance in emulation for some or all sources. To avoid making such a compromise, we can use the latent space to identify the dissimilar data source(s) and then repeat the fusion process after excluding them.

Note also that the objective function in Eq. **8** that is used to find the latent positions is invariant under translation and rotation. In order to find a unique solution, we enforce the following constraints in two dimensions (more constraints are needed for $d_z > 2$): latent point 1 is placed at the origin, latent point 2 is positioned on the positive $x$ axis, and latent point 3 is restricted to the $y > 0$ half-plane. We assign $y_h(x)$ to position 1 for both of our strategies as it yields more readable latent plots, but this choice is arbitrary and does not affect the relative distances between the latent positions as shown in Section 4.

Returning to our example with the above constraints in mind, we can see that the latent points corresponding to $y_h(x)$ and $y_{l_2}(x)$ are close and the other points relatively distant, especially the point representing $y_{l_3}(x)$. This observation matches with our knowledge on the relative accuracies of the underlying functions with respect to $y_h(x)$ (this knowledge is *not* provided to LMGP). In other words, LMGP has accurately determined the correlations between the data sources despite the sparse sampling for $y_h(x)$. Given that $y_{l_2}(x)$ appears to be much more accurate than other low-fidelity sources with respect to $y_h(x)$, one might consider fitting LMGP using only data from these two sources rather than all of the data to produce a more accurate high-fidelity emulator. The results of this approach, shown in **Figure 2 (d)**, demonstrate that high-fidelity emulation performance is actually equivalent with all sources used, i.e., using less accurate sources does not make our estimate of $y_h(x)$ worse in this case because they include useful information about $y_h(x)$.

Consider now a second pedagogical example with three datasets drawn from the following functions:

$$y_h(x) = 0.1x^3 + x^2 + x + 1, \quad -2 \leq x \leq 3 \qquad \text{Eq. 16.1}$$

$$y_{l_1}(x) = 0.2x^3 + x^2 + x + 1, \quad -2 \leq x \leq 3 \qquad \text{Eq. 16.2}$$

$$y_{l_2}(x) = x^2 + x + 1, \quad -2 \leq x \leq 3 \qquad \text{Eq. 16.3}$$

where we again sample via Sobol sequence with $n_h = 3$, $n_{l_1} = n_{l_2} = 20$, and do not apply noise to the samples. We create 30 unique quasi random iterations (hereafter referred to as repetitions) to examine the robustness of our approach to sampling variations. As shown in **Figure 3 (a)**, both $y_{l_2}(x)$ and $y_{l_1}(x)$ are equally accurate as they differ from $y_h(x)$ by a $\pm 0.1x^3$ term. This time, we fit LMGP using both strategies for categorical variable assignment and examine the effect of this choice as well as the size of the training datasets on the results. We use the subscript *All* to denote the fact that we fit LMGP to all available data and employ the subscripts $l_i$ to refer to an LMGP fitted via only $\boldsymbol{y}_h$ and $\boldsymbol{y}_{l_i}$.

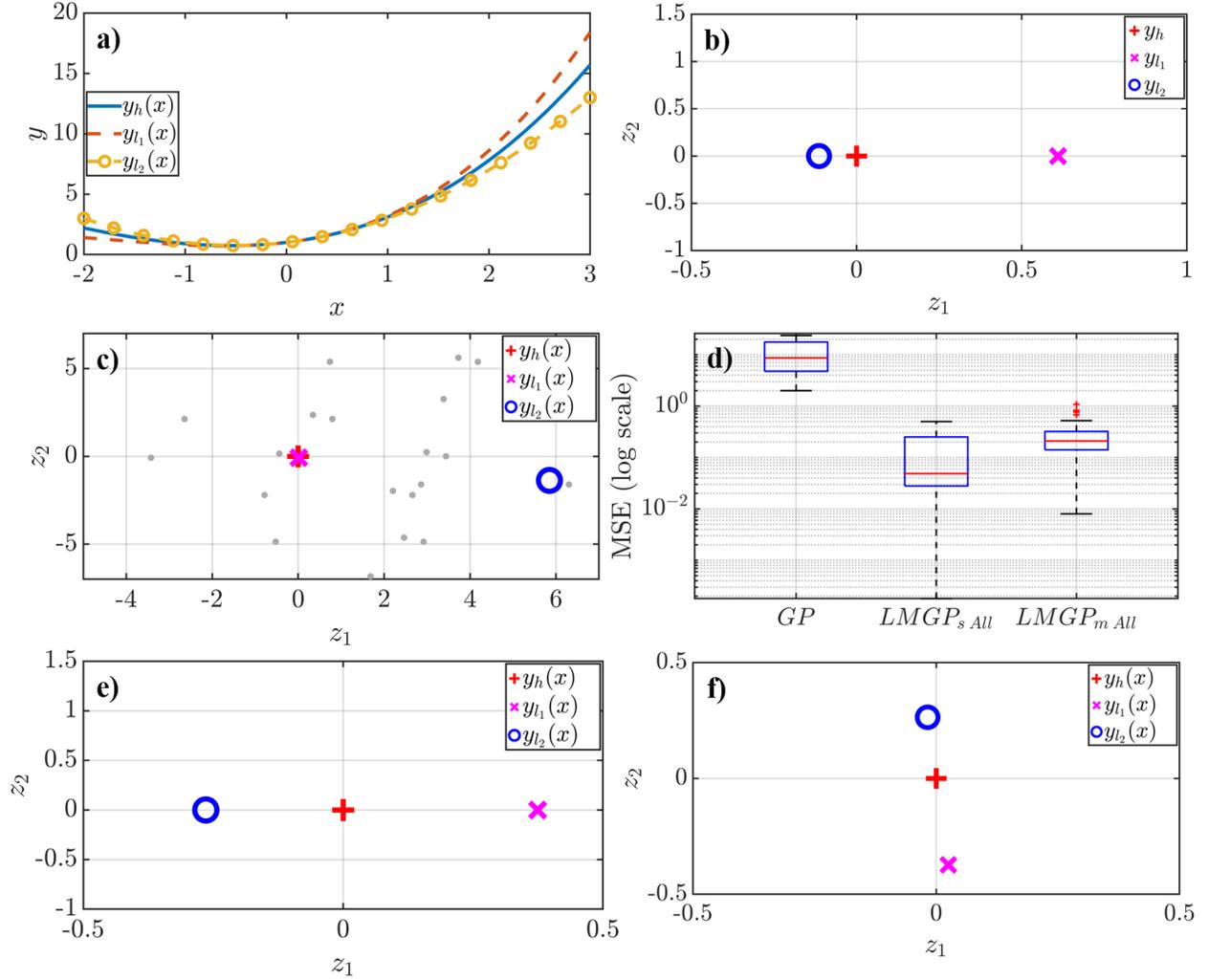

**Figure 3 Approaches to categorical variable assignment: (a) Accuracy of data sources:** Both low-fidelity sources are equally accurate. **(b) Latent space for LMGP$_{s\ All}$:** We show the latent space for one repetition, but LMGP consistently finds one source to be close to and another to be distant from the position for $y_h(x)$ across repetitions. **(c) Latent space for LMGP$_{m\ All}$:** We show the latent space for one repetition. The positions and relative distances are not consistent across repetitions. The gray dots correspond to latent positions that do not correspond to any data source. **(d) High-fidelity emulation performance across 30 repetitions:** LMGP outperforms GP in high-fidelity emulation for both categorical variable strategies. MSEs are calculated by comparing emulator predictions to analytic function outputs at 10,000 points. **(e) and (f) Latent spaces with more data:** With more data, LMGP$_{s\ All}$, shown in **(e)**, and LMGP$_{m\ All}$, shown in **(f)**, consistently find latent positions that accurately reflect the relative accuracies of the data sources. We do not show the latent positions not corresponding to any data sources in **(f)**.

The latent space for LMGP using one categorical variable is demonstrated in **Figure 3 (b)** and shows that this strategy enables LMGP to learn that both sources have inaccuracy with respect to $y_h(x)$. However, LMGP consistently finds one source to be significantly more accurate than the other as a result of the sparse sampling. By contrast, the positions found by LMGP using multiple categorical variables are very inconsistent across repetitions and often estimate one of the sources as being either extremely correlated or uncorrelated with $y_h(x)$, see **Figure 3 (c)**. This inconsistency is due to the fact that LMGP$_{m\ All}$ has quite a few hyperparameters (1 roughness parameter and 18 parameters in the $\boldsymbol{A}$ matrix) which are difficult to estimate with scarce data. Across the repetitions of LMGP$_{m\ All}$, at least one data source is always found to be well correlated with $y_h(x)$ so high-fidelity predictions are still good and much better than fitting a traditional GP to only the high-fidelity data, see **Figure 3 (d)**. When we increase the available data to $n_h = 15$, $n_{l_1} = n_{l_2} = 50$, both LMGP$_{s\ All}$ and LMGP$_{m\ All}$ consistently (i.e., across repetitions) find latent positions for the low-fidelity functions that are approximately equidistant from $y_h(x)$, see **Figure 3 (e)** and **(f)**. Interestingly, these positions are in opposite directions which agrees with the fact that discrepancies are equal but of opposite sign.

While we did not apply noise to the samples in these pedagogical examples, as we demonstrate in Section 4 LMGP is fairly robust to noise both with respect to emulation performance and finding latent positions.

## 3.2 Calibration via LMGP

Calibration problems closely resemble multi-fidelity modeling in that a number of high- and low-fidelity datasets are assimilated or fused together. However, in such problems low-fidelity datasets[3] typically involve calibration inputs which are not directly controlled, observed, or measured in the high-fidelity data (i.e., high-fidelity data have fewer inputs). Hence, in addition to building surrogate models, one seeks to *inversely* estimate these inputs during the calibration process. The estimated calibration inputs either quantify the system's properties (e.g., estimating

---

[3] Generally built via computer simulations

Youngs modulus based on a tension test) or serve as tuning parameters that aim to compensate for the missing physics (aka model form error) in simulators. Knowledge of a simulator's correct calibration parameters for a given dataset is highly desirable as it allows to use the simulator for cases where obtaining high-fidelity data is expensive or infeasible.

Following previous sections, we denote the quantitative and latent representation of the qualitative inputs via $x$ and $z$, respectively (note that $z$ encode data sources as per Section 3.1). While these inputs are shared across all data sources, the low-fidelity data sources have additional quantitative inputs, $\boldsymbol{\theta}$, whose "best" values must be estimated using the high-fidelity data. As in section 2.2, we represent these "best" values by $\boldsymbol{\theta}^*$ which minimize the discrepancies between low- and high-fidelity datasets based on an appropriate metric. In the case that one wishes to calibrate and assimilate multiple computer models simultaneously, we assume that the calibration parameters are shared between the low-fidelity datasets and are expected to have the same best value. Our estimate of $\boldsymbol{\theta}^*$ is denoted by $\widehat{\boldsymbol{\theta}}$ and is obtained via MLE by modifying LMGP's correlation function as:

$$
\begin{aligned}
r\left(\begin{bmatrix} x \\ z \\ \boldsymbol{\theta} \end{bmatrix}^{(i)}, \begin{bmatrix} x \\ z \\ \boldsymbol{\theta} \end{bmatrix}^{(j)}\right) \\
= exp\left\{-\left(z^{(i)} - z^{(j)}\right)^T\left(z^{(i)} - z^{(j)}\right)\right\} \\
\times exp\left\{-\left(x^{(i)} - x^{(j)}\right)^T \boldsymbol{\Omega}_x \left(x^{(i)} - x^{(j)}\right)\right\} \\
\times exp\left\{-\left(\boldsymbol{\theta}^{(i)} - \boldsymbol{\theta}^{(j)}\right)^T \boldsymbol{\Omega}_\theta \left(\boldsymbol{\theta}^{(i)} - \boldsymbol{\theta}^{(j)}\right)\right\}
\end{aligned}
\qquad \text{Eq. 17}
$$

where $x^{(i)}, x^{(j)}, \boldsymbol{\Omega}_x, z^{(i)}$, and $z^{(j)}$ are defined as before. $\boldsymbol{\theta}^{(i)}$ denote the calibration parameters of sample $i$ and $\boldsymbol{\Omega}_\theta$ is the diagonal matrix of roughness/scale parameters associated with $\boldsymbol{\theta}$. When one or both of the inputs to the correlation function lack calibration parameters (i.e., at least one of the inputs corresponds to a high-fidelity sample), we substitute $\widehat{\boldsymbol{\theta}}$ in the last term of Eq. 17. If both inputs are from the high-fidelity data, the term $exp\left\{-\left(\boldsymbol{\theta}^{(i)} - \boldsymbol{\theta}^{(j)}\right)^T \boldsymbol{\Omega}_\theta \left(\boldsymbol{\theta}^{(i)} - \boldsymbol{\theta}^{(j)}\right)\right\}$ does not affect the correlation because $exp\left\{-\left(\widehat{\boldsymbol{\theta}} - \widehat{\boldsymbol{\theta}}\right)^T \boldsymbol{\Omega}_\theta \left(\widehat{\boldsymbol{\theta}} - \widehat{\boldsymbol{\theta}}\right)\right\} = exp\{0\} = 1$.

Using Eq. 17 we see that in a calibration problem with multiple data sources all the hyperparameters of an LMGP can be estimated by MLE in the same way that a traditional GP is trained, i.e., by optimizing the following objective function where the correlation matrix is built via Eq. 17:

$$[\widehat{\omega}, \widehat{A}, \widehat{\theta}, \widehat{\Omega}_\theta] = \underset{\omega, A, \theta, \Omega_\theta}{\text{argmin}} \ nlog(\widehat{\sigma}^2) + log(|R|) = \underset{\omega, A, \theta, \Omega_\theta}{\text{argmin}} L,$$   Eq. 18

Preprocessing the data for calibration via LMGP is schematically illustrated in **Figure 4**. Following the same procedure described in Section 3.1 we append the inputs with categorical variables to distinguish data sources. We also augment the high-fidelity inputs with some unknown values to account for the missing calibration parameters. Once the mixed dataset that contains *all* the low- and high-fidelity data is built, we directly use it in LMGP to not only build emulators for each data source, but also estimate $\widehat{\theta}$. Similar to multi-fidelity modeling, any number of datasets can be simultaneously used via LMGP for calibration.

We now illustrate the capabilities of LMGPs for calibration via two analytical examples where there are one high-fidelity data source $y_h(x)$ and up to two low-fidelity data sources, denoted by $y_{l_1}(x)$ and $y_{l_2}(x)$. We presume that in both examples the goals are to accurately emulate the high-fidelity data source and estimate the calibration parameters. We note that once an LMGP is trained, it provides an emulator for each data source but here we only evaluate accuracy for $y_h(x)$ since much fewer data points are available from it and hence emulating it is more difficult. For our first example, we consider the polynomials in Eq. 19 as data sources and take 5 samples from $y_h(x)$ and 25 samples from each of $y_{l_1}(x)$ and $y_{l_2}(x)$ (none of the datasets are corrupted with noise):

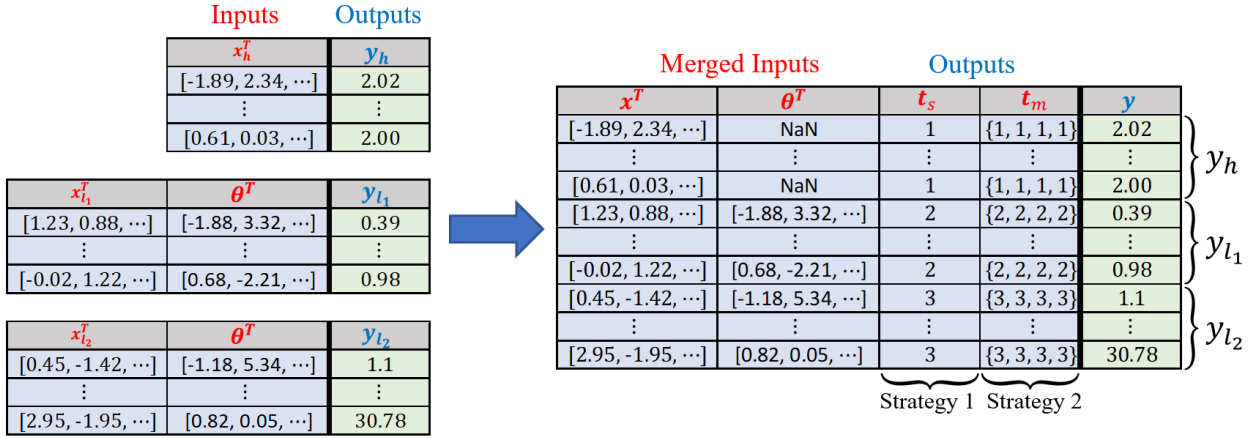

**Figure 4 Pre-processing of data for calibration:** Multiple datasets are combined in a specific way and then directly used by LMGP. The high-fidelity data are augmented with NaNs since they lack calibration parameters, and all data are augmented with categorical IDs that denote the source from which a datum is drawn. We use strategy 1 for choice of $t$ in both examples in section 3.2.

$$y_h(x) = 0.1x^3 + x^2 + x + 1, \quad -2 \leq x \leq 3 \qquad \text{Eq. 19.1}$$

$$y_{l_1}(x) = \theta x^3 + x^2 + x + 1, \quad -2 \leq x \leq 3, \quad -2 \leq \theta \leq 2 \qquad \text{Eq. 19.2}$$

$$y_{l_2}(x) = \theta x^3 + x^2 + 1, \quad -2 \leq x \leq 3, \quad -2 \leq \theta \leq 2 \qquad \text{Eq. 19.3}$$

We set $\theta^*$ as 0.1 because it is the true value of the coefficient on the leading $x^3$ term. Note that $y_{l_1}(x)$ can match $y_h(x)$ perfectly with appropriate choice of $\theta$, i.e., $y_{l_1}(x)$ has no model form error when $\hat{\theta} = 0.1$, see **Figure 5 (a)**. Conversely, no value of $\theta$ allows $y_{l_2}(x)$ to match $y_h(x)$ since $y_{l_2}(x)$ has a linear model form error. When solving this calibration problem, we assume there is no knowledge on whether low-fidelity models have discrepancies and expect the learned latent space of LMGP to provide diagnostic measures that indicate potential model form errors.

As shown in **Figure 5 (b)**, the learned latent positions by LMGP are quite consistent with our expectations despite the fact that limited and unbalanced data are used in LMGP's training. It is evident that the latent positions corresponding to $y_h(x)$ and $y_{l_1}(x)$ are very close to each other, indicating negligible model form error. In contrast, the positions corresponding to $y_h(x)$ and $y_{l_2}(x)$ are more distant which signals that $y_{l_2}(x)$ has model form error.

The learned latent positions in **Figure 5 (b)** suggest that $y_{l_1}(x)$ (when calibrated properly) captures the behavior of $y_h(x)$ better than $y_{l_2}(x)$. Correspondingly, one may argue calibrating $y_{l_1}(x)$ individually may improve performance. To assess this argument, we fit LMGPs to three combinations of the available datasets and compare the performance of these LMGPs in terms of estimating $\theta^*$ and emulating $y_h(x)$. In all three cases, we use a single categorical variable to encode the data source and hence the subscript $s$ is appended to the model names (so, LMGP$_{s\,l_1}$ calibrates $y_{l_1}(x)$ via $y_h(x)$ and uses a single categorical variable). The results are shown in **Figure 5 (c)** and **Figure 5 (d)** and indicate that using both low-fidelity datasets provides the best performance since $(i)$ $\hat{\theta}$s are estimated more consistently as the distribution is centered at $\theta^*$ with small variations, and $(ii)$ errors (measured in terms of mean squared error, MSE) for predicting $y_h(x)$ are smaller. These observations can be explained by the fact that the highest relative distance between data sources in **Figure 5 (b)** is on the order of 0.5 which indicate that $y_{l_2}(x)$ is somewhat similar to $y_h(x)$ and $y_{l_1}(x)$ as this distance scales the correlation function by $\exp\{(-0.5)^2\} \approx$ $0.78 \gg 0$. That is, LMGP can distill useful knowledge from the weak correlation between $y_{l_2}$ and other sources to improve its performance in estimating $\theta$ and emulating $y_h(x)$. When $y_{l_1}(x)$ is excluded from the calibration process and only $y_{l_2}(x)$ is used in calibration, LMGP provides biased and less consistent estimates for $\theta$ and relatively large MSEs for predicting $y_h(x)$.

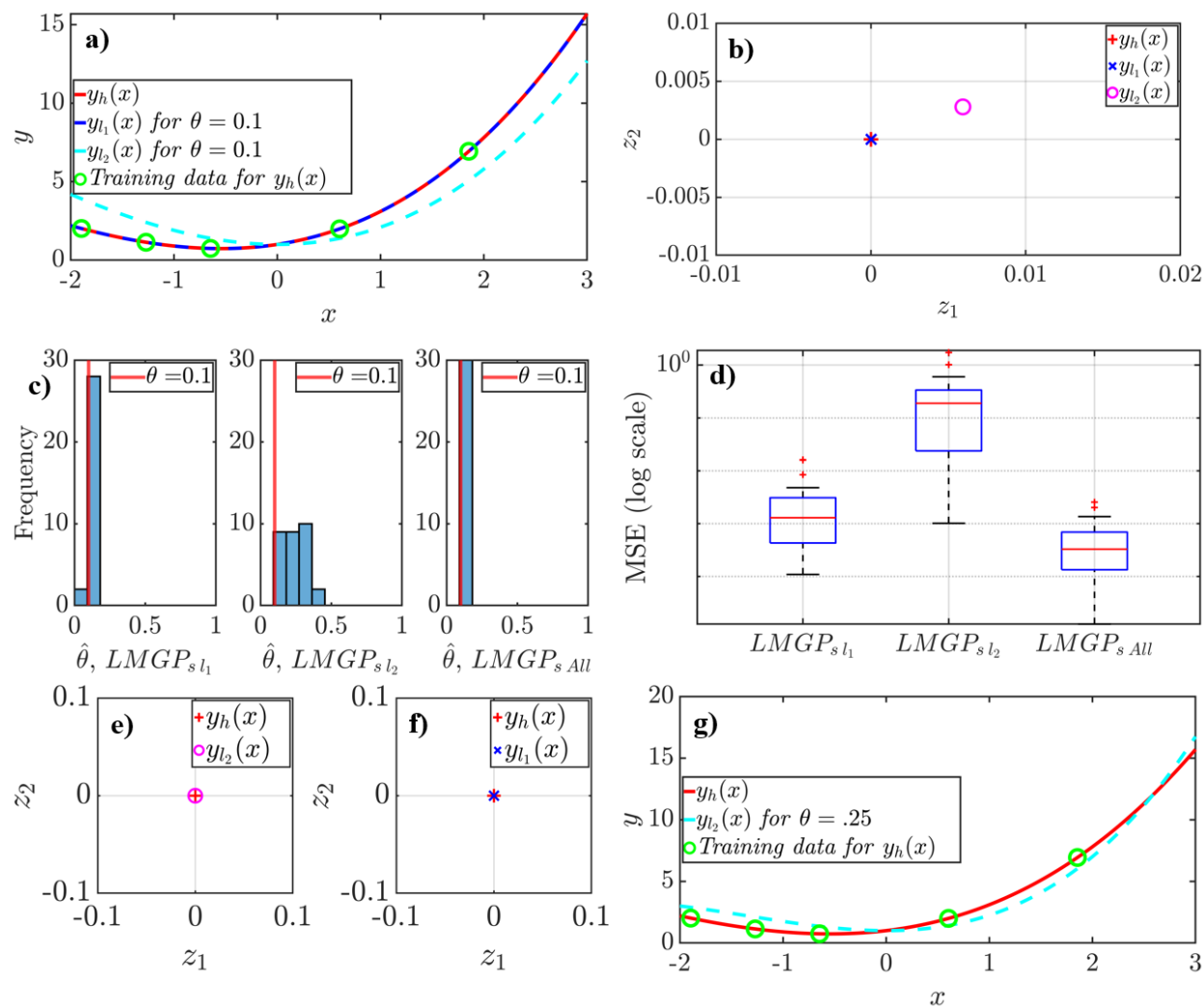

**Figure 5 Calibration with LMGP: (a) Underlying functions with true calibration parameters:** $y_{l_1}(x)$ and $y_h(x)$ are coincident for $\theta = \theta^*$. **(b) Latent space for LMGP$_{s\,All}$:** Latent positions for $y_h(x)$ and $y_{l_1}(x)$ are coincident while the position for $y_{l_2}(x)$ is relatively more distant (albeit still quite close). **(c) Histogram of estimated calibration parameters:** We estimate $\theta$ over 30 repetitions where the LMGP fitted via all data yields more consistent estimates. All three models use a single categorical variable to encode data sources. **(d) High-fidelity emulation performance:** Using all data yields the best performance since data sources are correlated. **(e) and (f) Latent space for LMGP$_{s\,l_2}$ and LMGP$_{s\,l_1}$:** LMGP cannot detect model form error between $y_h(x)$ and $y_{l_2}(x)$ since data is scarce and an appropriately estimated $\theta$ enables $y_{l_2}(x)$ to resemble $y_h(x)$ fairly well. LMGP can correctly detect that $y_{l_1}(x)$ does not have model form error. **(g)** $y_{l_2}(x)$ **with estimated calibration parameters versus** $y_h(x)$: $y_{l_2}(x)$ can nearly interpolate sparse training data for $y_h(x)$ with the appropriate calibration parameter.

While the distance in the latent space typically encodes model form error that is not reduceable by adjusting $\theta$, LMGP may mistake model form error for noise in the case that certain calibration parameters allow the low-fidelity model to closely match the high-fidelity function. This is the case if we fit LMGP to only $y_h(x)$ and $y_{l_2}(x)$. As shown in **Figure 5 (e)**, LMGP places the latent positions for $y_h(x)$ and $y_{l_2}(x)$ very close to each other when $y_{l_1}(x)$ is excluded. We explain this observation by referring back to **Figure 5 (c)** where LMGP$_{s\ l_2}$ finds $\hat{\theta} \approx 0.25$. Plotting $y_{l_2}(x)$ for this value of $\theta$ reveals that it can nearly interpolate the training data, see **Figure 5 (g)**. As such, LMGP mistakes 0.25 for the true value of $\theta$ and dismisses the small resultant error as noise. This also explains the aforementioned bias and inconsistency in estimating $\theta$ across repetitions as the value that comes closest to interpolating $y_h(x)$ is different depending on sampling variations. By contrast, LMGP fit to all data is able to leverage the information from $y_{l_1}(x)$ to determine that $y_{l_2}(x)$ has model form error. And, as expected, no model form error is indicated in the latent space if only $y_{l_1}(x)$ is used in calibration, see **Figure 5 (f)**. As this simple example clearly indicates, simultaneous fusion of *multiple* (i.e., more than 2) data sources can decrease identifiability issues in calibration. This property is one of the main strengths of our data fusion approach.

In our second analytical example we examine a case where there is only one low-fidelity source which has a significant model form error:

$$y_h(x) = \sin(\pi x) + \sin(10\pi x), \qquad 0 \leq x \leq 1 \qquad \text{Eq. 20.1}$$

$$y_l(x) = \sin(\theta x), \qquad 0 \leq x \leq 1 \text{ and } \pi - 2 \leq \theta \leq 10\pi + 2 \qquad \text{Eq. 20.2}$$

Based on Eq. 20, $\theta^*$ can be either $\pi$ or $10\pi$ so the range of $\theta$ in $y_l(x)$ is chosen wide enough to include both values. As shown in **Figure 6 (a)**, considering $\theta^* = \pi$ implies that the high-fidelity source is either noisy or has a high frequency component that is missing from the low-fidelity source (note that in realistic applications the functional form of data sources is unknown so high frequency trends can be easily misclassified as noise in which case they are typically smoothed out, i.e., not learned). Conversely, considering $\theta^* = 10\pi$ implies that $y_l(x)$ is expected to surrogate the high-frequency component of $y_h(x)$ and that $\sin(\pi x)$ is the discrepancy. Note that the analytic MSEs (calculated by comparing $y_h(x)$ and $y_l(x)$ at 10,000 sample points equally spaced over the input range) and cosine similarities (between $y_h(x)$ and $y_l(x)$, also at 10,000 sample points equally spaced over the input range) are identical for each choice of $\theta$, i.e., both choices yield a discrepancy of the same magnitude and we cannot determine which choice is better *a priori* based on MSEs or cosine similarity. We are interested in finding out which value is a better estimate for $\theta^*$ and whether LMGP is able to consistently infer this value purely from the low- and high-fidelity datasets. We do not corrupt the datasets with noise and investigate the effect of noise in Section 4.2.

We now explore the effects of the low-fidelity dataset size on the performance while holding the number of high-fidelity data constant. Specifically, we examine $n_l = 30, 100, 200$ with $n_h = 15$ in each case. Note that standard GP trained on only the 15 available high-fidelity samples cannot learn the high-frequency behavior of $y_h(x)$ and instead interprets it as noise.

As shown in **Figure 6 (b)**, increasing $n_l$ improves high-fidelity prediction and we can therefore consider the estimates of $\theta$ and the latent distances in the $n_l = 200$ case to be the most accurate since they maximize prediction performance. Shown in **Figure 7 (a)** are histograms of the latent distances over 30 repetitions for each case. When few low-fidelity data are available, the latent distances are close to zero; with plentiful data, the latent distances are clustered around 0.5. This indicates that LMGP interprets $y_h(x)$ and $y_l(x)$ as being closely correlated when we have few low-fidelity data, but consistently learns that $y_l(x)$ has noticeable error with respect to $y_h(x)$ as we provide more data. Without sufficient low-fidelity data, LMGP learns the low-frequency behavior of $y_h(x)$ which follows $\sin(\pi x)$ and dismisses the high-frequency behavior as noise. Consequently, LMGP finds a small latent distance since $y_l(x)$ can capture $\sin(\pi x)$ without error.

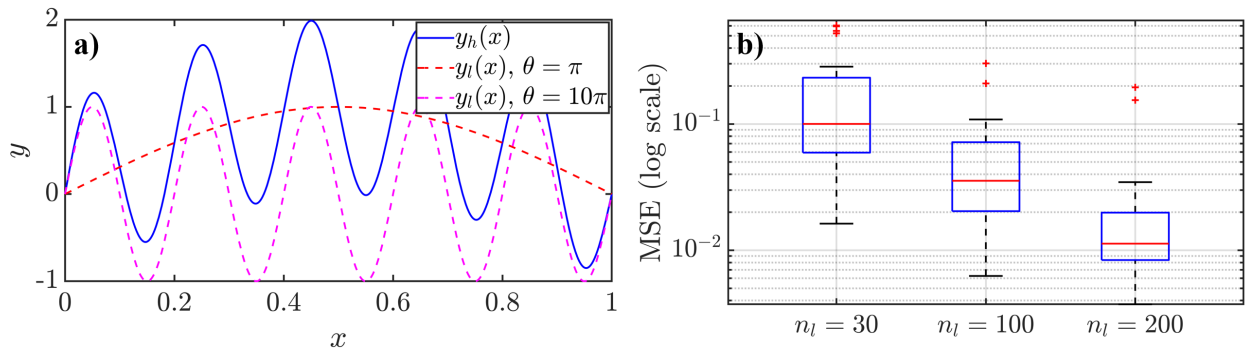

**Figure 6 Calibration via LMGP: (a) Plot of the underlying functions:** Due to model form error, $y_l(x)$ is unable to capture the behavior of $y_h(x)$ regardless of the choice of $\theta$. Choosing $\theta = \pi$ indicates a discrepancy of $\sin(10\pi)$, while choosing $\theta = 10\pi$ indicates a discrepancy of $\sin(\pi)$. Notably, the analytic MSEs (calculated by comparing $y_h(x)$ and $y_l(x)$ at 10,000 sample points equally spaced over the input range) for both choices of theta are 0.5, i.e., the magnitude of the error is the same for both choices of $\theta$. **(b) High fidelity emulation performance:** As we provide more low-fidelity data, LMGP's performance on high-fidelity emulation increases.

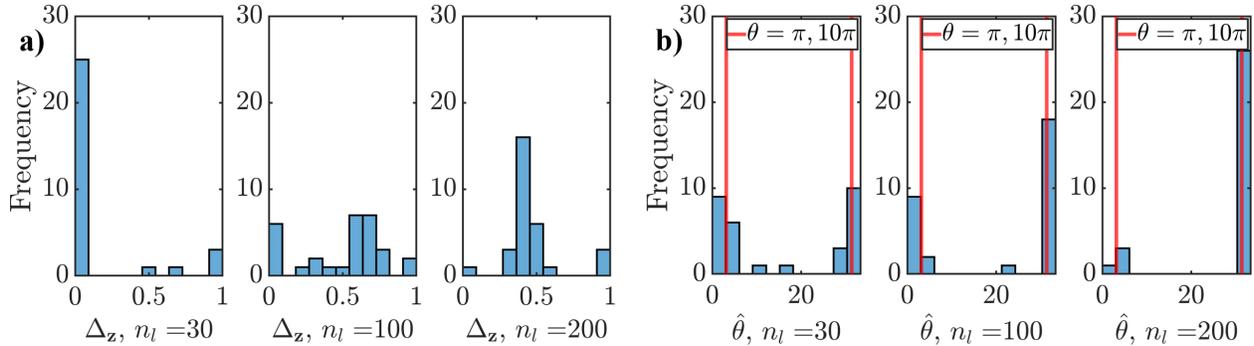

**Figure 7 Analysis for sin wave example: (a) Histogram of latent distances:** LMGP estimates distances near zero and 0.5 with a few and plentiful data points, respectively. There is large variance in the latent distances for $n_l = 100$, with a large spike at zero and a cluster near 0.5 which correspond to LMGP's estimates for $n_l = 30$ and $n_l = 200$ respectively. That is, as the size of the data is increasing, LMGPs interpretation of model form error changes. **(b) Histogram of $\hat{\theta}$:** As more low-fidelity data are provided, estimates become more closely clustered around $10\pi$. With few low-fidelity data, LMGP guesses $\theta = \pi$ almost half of the time but with $n_l = 200$ LMGP almost consistently guesses $\theta = 10\pi$ which means that $y_l(x)$ has a high-frequency behavior.

We now examine the histogram of $\hat{\theta}$ in **Figure 7 (b)**. When few low-fidelity data are available, estimates are clustered around both $\pi$ and $10\pi$ while with plentiful data the estimates are tightly clustered around only $10\pi$. This observation indicates that when little data is available, LMGP interprets $y_h(x)$ to more closely resemble $\sin(\pi x)$ almost half of the times which matches with the observation on the learned latent distances, i.e., the high-frequency behavior is interpreted as noise and not learned. As more low-fidelity data is available, LMGP is able to learn the high-frequency behavior of $y_h(x)$ using the low-fidelity data and interprets $y_h(x)$ as more closely resembling $\sin(10\pi x)$.

Why does LMGP prefer $\hat{\theta} = 10\pi$ with more data? To answer this question, we note that in LMGP shifting the levels of the categorical variable is expected to reflect a change in data source. With $\hat{\theta} = \pi$ the shift in the categorical variable is supposed to "model" $\sin(10\pi x)$ which is much more difficult than the alternative. In other words, LMGP is trying to learn the simplest function that must be represented by a shift in the categorical variable, see **Figure 8 (a)**. We further explore this conjecture by fitting an LMGP to 100 noiseless samples from $y_h(x)$ and 200 samples from $y_l(x)$. This amount of data is sufficient to learn both the high-frequency behavior of $y_h(x)$ and the high-frequencies of $y_l(x)$ (i.e., the behavior of $y_l(x)$ for large $\theta$), and as such we expect the latent positions and calibration estimates found by LMGP in this case to be optimal. As shown in **Figure 8 (b)**, LMGP finds latent distances near 0.5 and $\theta = 10\pi$ very consistently, i.e., LMGP prefers to estimate the calibration parameters to minimize the complexity of the discrepancy function.

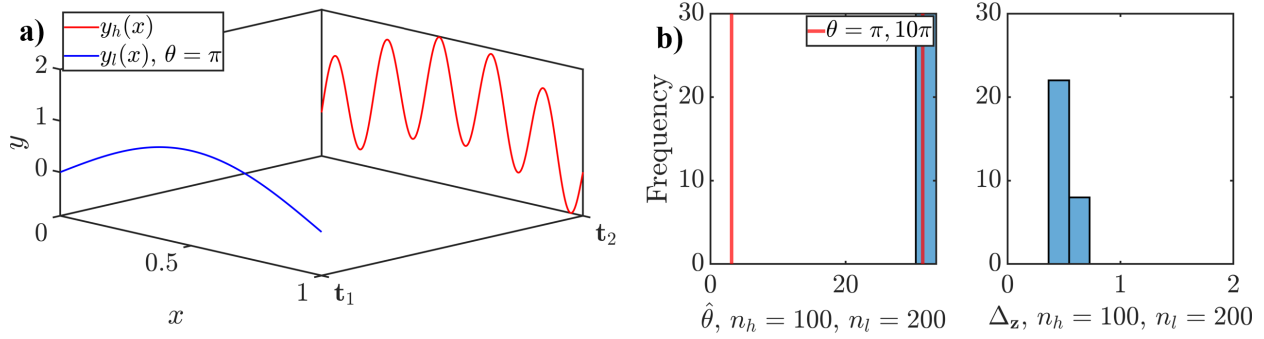

**Figure 8 Effect of categorical variable and dataset size: (a) Effect of shifting the level:** With $\hat{\theta} = \pi$ the shift in the categorical variable is supposed to "model" $\sin(10\pi x)$ which is much more difficult than the alternative. **(b) Effect of dataset size:** with $n_h = 100$ and $n_l = 200$ LMGP consistently estimates $\theta$ as $10\pi$ so the shift in categorical variable learns the simplest discrepancy candidate, i.e., $\sin(\pi x)$.

## 4 Results

To validate our approach in both multi-fidelity and calibration problems, we test our method on analytical functions and assess its performance against competing methods. In each example, we vary the size of the training data and added noise variance and repeat the training process 30 times to account for randomness (the knowledge on the value of the noise variance is *not* used in training). To measure accuracy, we use 10,000 noisy test samples to obtain MSE (note that since the test data are noisy, the MSE obtained by an emulator cannot be smaller than the noise variance).

In our LMGP implementation, we always use $d_z = 2$ and select $-3 \leq a_{i,j} \leq 3$ during optimization where $a_{i,j}$ are the elements of the mapping matrix $\boldsymbol{A}$. When using LMGP for calibration, the search space for each element of $\hat{\boldsymbol{\theta}}$ is restricted to $[-2, 3]$ after scaling the data to the range $[0, 1]$ (i.e., we select a search space larger than the sampling range for $\boldsymbol{\theta}$). We use the modular version of KOH's approach where we set a uniform prior for $\boldsymbol{\theta}$ over the sampling range defined in each problem statement.

### 4.1 Multi-Fidelity Results

We consider two analytical problems with high dimensional inputs. In the first multi-fidelity problem, we consider a set of four functions that model the weight of a light aircraft wing [48]:

$$y_h(x) = 0.036 S_\omega^{0.758} W_{f\omega}^{0.0035} \left(\frac{A}{\cos^2(\Lambda)}\right)^{0.6} q^{0.006} \lambda^{0.04} \left(\frac{100 t_c}{\cos(\Lambda)}\right)^{-0.3} (N_z W_{dg})^{0.49} + S_\omega W_p \qquad \text{Eq. 21.1}$$

$$y_{l_1}(x) = 0.036 S_\omega^{0.758} W_{f\omega}^{0.0035} \left(\frac{A}{\cos^2(\Lambda)}\right)^{0.6} q^{0.006} \lambda^{0.04} \left(\frac{100 t_c}{\cos(\Lambda)}\right)^{-0.3} (N_z W_{dg})^{0.49} + 1 \times W_p \qquad \text{Eq. 21.2}$$

$$y_{l_2}(x) = 0.036 S_\omega^{0.8} W_{f\omega}^{0.0035} \left(\frac{A}{\cos^2(\Lambda)}\right)^{0.6} q^{0.006} \lambda^{0.04} \left(\frac{100 t_c}{\cos(\Lambda)}\right)^{-0.3} (N_z W_{dg})^{0.49} + 1 \times W_p \qquad \text{Eq. 21.3}$$

$$y_{l_3}(x) = 0.036 S_\omega^{0.9} W_{f\omega}^{0.0035} \left(\frac{A}{\cos^2(\Lambda)}\right)^{0.6} q^{0.006} \lambda^{0.04} \left(\frac{100 t_c}{\cos(\Lambda)}\right)^{-0.3} (N_z W_{dg})^{0.49} + 0 \times W_p \qquad \text{Eq. 21.4}$$

$$x^T = [S_\omega, W_{f\omega}, A, \Lambda, q, \lambda, t_c, N_z, W_{dg}, W_p]$$

$$\min(x) = [150, 220, 6, -10, 16, 0.5, 0.08, 2.5, 1700, 0.025]$$

$$\max(x) = [200, 300, 10, 10, 45, 1, 0.18, 6, 2500, 0.08]$$

These functions are ten-dimensional and have varying degrees of fidelity where, following the notation introduced in Sec. 3, $y_h(x)$ has the highest fidelity. Note that in $y_{l_3}(x)$, we multiply $W_p$ by zero which is equivalent to reducing the dimensionality of the function by one. As enumerated in **Table 2**, the above functions are listed in decreasing order with respect to accuracy, i.e., $y_{l_1}(x)$ and $y_{l_3}(x)$ are the most and least accurate models, respectively. **Table 2** is generated by evaluating the four functions in Eq. 21 on the same 10,000 inputs as described in section 3.1 (no noise is added to the outputs). This knowledge on relative accuracy of the of data sources is *not* used when fitting an LMGP.

**Table 2 Relative accuracy of functions for wing weight problem:** The functions are listed in decreasing order with respect to accuracy, with $y_{l_3}(x)$ being especially inaccurate. 10000 points are used in calculating RRMSE.

|  | $y_{l_1}(x)$ | $y_{l_2}(x)$ | $y_{l_3}(x)$ |
|---|---|---|---|
| **RRMSE** | 0.19912 | 1.1423 | 5.7484 |

We consider various amounts of available low-fidelity data, with and without noise. We also compare the two different settings introduced in Sec. 3.1 where subscripts $s$ and $m$ indicate whether a single or multiple categorical variables are used to encode the data sources in LMGP. We only take 15 samples for $y_h(x)$ which is a very small number given the high dimensionality of the input space. Additionally, we investigate the effect of fusing the four datasets jointly against fusing the high-fidelity data with the most accurate low-fidelity source which is $y_{l_1}(x)$ (in the former case the subscript *All* is appended to LMGP while in the latter case $l_1$ is used in the subscript).

The results are summarized in **Figure 9** and indicate that the different versions of LMGPs consistently outperform traditional GPs (only fitted to high-fidelity data) in all cases. This superior performance of LMGP is due to taking advantage of the correlations between datasets that compensates, to some extent, for the sparsity of the high-fidelity data. LMGP also has the advantage in consistency where fewer outliers are observed in MSE compared to GP. This consistency indicates that our modeling assumptions (e.g., how to encode the data source) marginally affect the performance in this example.

In cases without noise, i.e., **Figure 9 (a)** and **(c)**, LMGPs fit to the data from $y_{l_1}(x)$ and $y_h(x)$ perform on par with or better than the LMGPs that are fit to all data and the small differences are mostly due to sample-to-sample variations. However, in cases with noise, i.e., **Figure 9 (b)** and **(d)**, using all the datasets improves the performance of LMGP. We explain this observation as follows: In the noiseless cases LMGP is able to quite accurately learn the behavior of $y_h(x)$ using just $y_{l_1}(x)$ and using all four datasets provides no additional advantage in learning $y_h(x)$ while (1) requiring the estimation of additional hyperparameters (in the $A$ matrix), and (2) compromising the estimates of $\Omega_x$ to handle the discrepancies between the four sources. By contrast, in the cases with noise one source is insufficient for LMGP to reach the threshold in emulation accuracy (which equals the noise variance) for $y_h(x)$. Including additional data sources in these cases helps LMGP to differentiate noise from model form error.

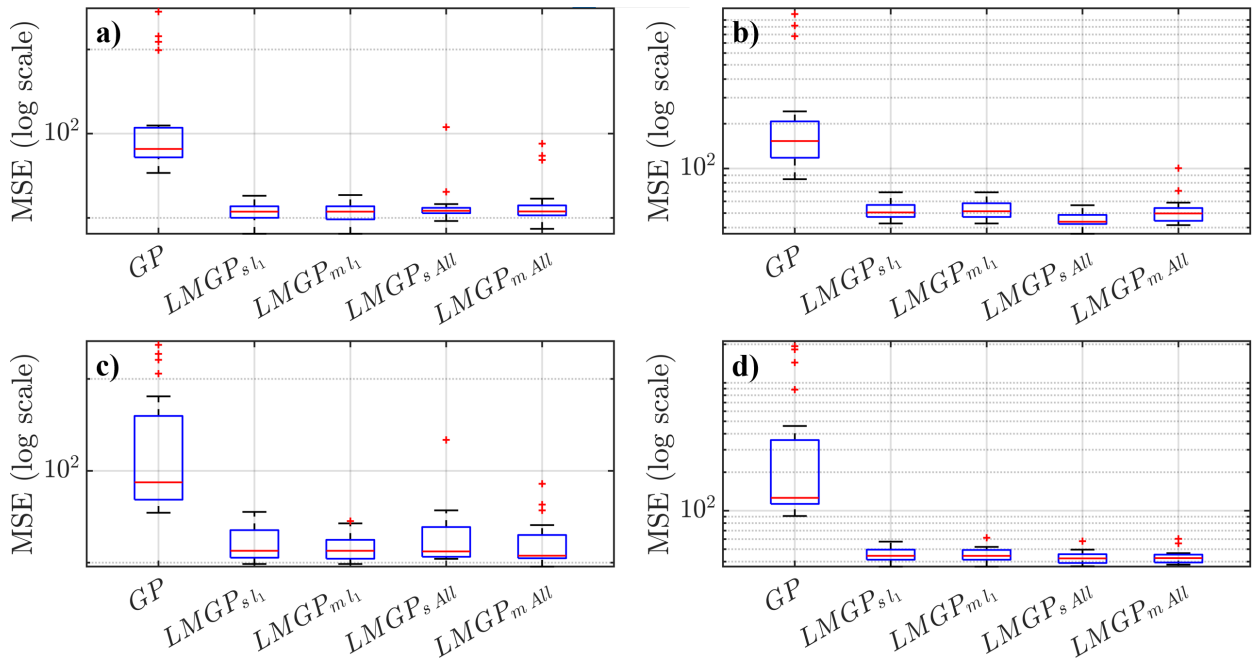

**Figure 9** High-fidelity emulation performance for wing weight example: (a) $n_h = 15$, $n_{l_1} = n_{l_2} = n_{l_3} = 50$, $\sigma^2 = 0$: All LMGP strategies perform at similar levels, with LMGP using only $y_{l_1}(x)$ arguably outperforming LMGP using all data sources. (b) $n_h = 15$, $n_{l_1} = n_{l_2} = n_{l_3} = 50$, $\sigma^2 = 25$: LMGP$_{s\ All}$ performs noticeably better than other LMGP strategies for this case. (c) $n_h = 15$, $n_{l_1} = n_{l_2} = n_{l_3} = 100$, $\sigma^2 = 0$: All LMGP strategies perform at similar levels, with LMGP using only $y_{l_1}(x)$ arguably outperforming LMGP using all data sources by a very slim margin. (d) $n_h = 15$, $n_{l_1} = n_{l_2} = n_{l_3} = 100$, $\sigma^2 = 25$: Both LMGP strategies that use all data sources outperform those that only use $y_{l_1}(x)$ and $y_h(x)$ by a slim margin.

For the remainder of this example, we investigate the most challenging version which has the fewest available data and highest level of noise. The latent space for this problem for LMGP$_{s\ All}$, shown in **Figure 10 (a)**, is once again a powerful diagnostic tool. While LMGP only has access to 15 noisy samples from the ten-dimensional function $y_h(x)$, the relative distances between latent positions match the relative accuracies of the data sources with respect to $y_h(x)$. The distance between $y_h(x)$ and $y_{l_3}(x)$ is $\approx 0.4$ yielding an approximate correlation of $\exp\{-(0.4^2)\} \approx 0.85$, which means that LMGP still uses information from $y_{l_3}(x)$ in predicting the response for $y_h(x)$ despite the former's low accuracy with respect to the latter.

We impose a number of constraints in order to obtain a unique solution for the latent positions since our objective function in Eq. **8** is invariant under translation and rotation. For a two-dimensional latent space, we fix the first position to the origin, the second position to the positive $z_1$ −axis, and the third position to the $z_2 > 0$ half-plane. As we mentioned before in section 3.1, we also assign the data sources to positions sequentially (i.e. $[y_h(x), y_{l_1}(x), y_{l_2}(x), y_{l_3}(x), \cdots] \rightarrow$

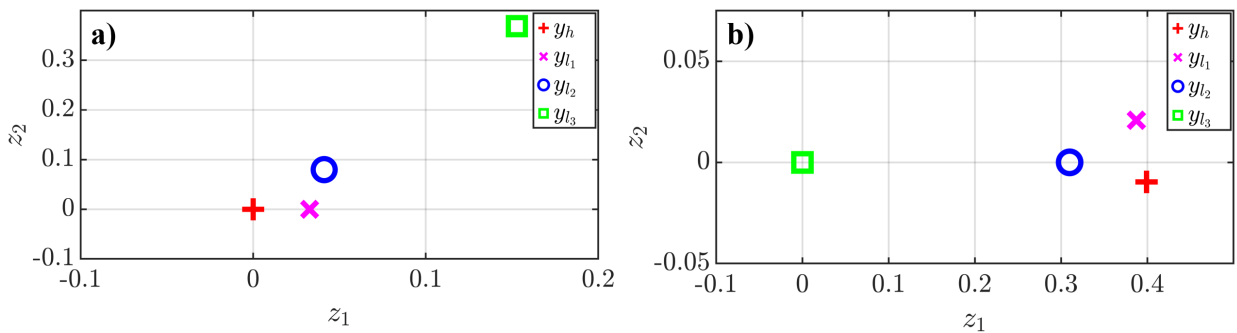

**Figure 10 Effect of constraints on the latent space: (a) Default constrains:** The latent space for one sample repetition of LMGP fit to all available data for the wing weight function with $n_h = 15$, $n_{l_1} = n_{l_2} = n_{l_3} = 50$, $\sigma^2 = 25$. $y_h(x), y_{l_1}(x)$, and $y_{l_2}(x)$ are positioned at, respectively, the origin, positive $z_1$ −axis, and first or second quadrant. While the learned latent spaces are different across the 30 repetitions, the relative latent distances are consistent both for different repetitions and for different amounts of data/noise. We only show the latent space of a randomly selected repetition. **(b) Alternate constraints:** The training procedure and data are exactly the same as before except that the three constrains are now applied to $y_{l_3}(x), y_{l_2}(x)$, and $y_{l_1}(x)$. Note that the relative distances between data sources is the same between the two plots.

[1, 2, 3, 4, ⋯]) with $y_h(x)$ at the origin for easier visualization of the relative correlations $y_{l_i}(x)$. While assigning the data sources to latent positions affects the learned latent positions, the relative distances between them remain the same as shown in **Figure 10 (b)**. Since we typically know the data source with the highest fidelity, the learned latent space of LMGP provides an extremely easy way to assess the fidelity of different sources with respect to it.

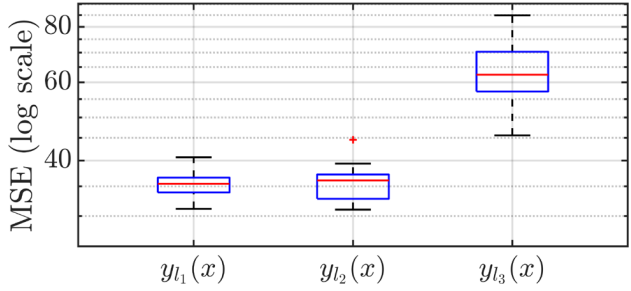

**Figure 11 Low-fidelity prediction performance:** Prediction accuracy is much higher for $y_{l_1}(x)$ and $y_{l_2}(x)$ than for $y_{l_3}(x)$.

Prediction performance on the low-fidelity sources for $LMGP_{s\ All}$, shown in **Figure 11**, follows the same trend as data source accuracy, i.e., it is best for $y_{l_1}(x)$ and worst for $y_{l_3}(x)$. When fitting LMGP to multiple data sources, we expect prediction accuracy to be high on sources that are well correlated with others, i.e., whose latent positions are close together or form a cluster. Leveraging information from a well-correlated source improves prediction performance more than the alternative, so each source in the cluster gains a boost in prediction performance from the information of the other sources in that cluster. In this case, $y_h(x)$, $y_{l_1}(x)$, and $y_{l_2}(x)$ form a cluster and as such we see that MSEs for $y_{l_1}(x)$ and $y_{l_2}(x)$ are much lower than those for $y_{l_3}(x)$.

In our next example, we consider data drawn from eight-dimensional models of water flow through a borehole [49]:

$$y_h(x) = \frac{2\pi T_u(H_u-H_l)}{ln\left(\frac{r}{r_w}\right)\left(1+\frac{2LT_u}{ln(r/r_w)r_w^2 K_w}+\frac{T_u}{T_l}\right)} \quad \text{Eq. 22.1}$$

$$y_{l_1}(x) = \frac{2\pi T_u(H_u-0.8H_l)}{ln\left(\frac{r}{r_w}\right)\left(1+\frac{1LT_u}{ln(r/r_w)r_w^2 K_w}+\frac{T_u}{T_l}\right)} \quad \text{Eq. 22.2}$$

$$y_{l_2}(x) = \frac{2\pi T_u(H_u-H_l)}{ln\left(\frac{r}{r_w}\right)\left(1+\frac{8LT_u}{ln(r/r_w)r_w^2 K_w}+0.75\frac{T_u}{T_l}\right)} \quad \text{Eq. 22.3}$$

$$y_{l_3}(x) = \frac{2\pi T_u(1.1H_u - H_l)}{ln\left(\frac{4r}{r_w}\right)\left(1 + \frac{2LT_u}{ln(r/r_w)r_w^2 K_w} + \frac{T_u}{T_l}\right)} \quad \text{Eq. 22.4}$$

$$x^T = [T_u, H_u, H_l, r, r_w, T_l, L, K_w]$$

$$\min(x) = [100, 990, 700, 100, 0.05, 10, 1000, 6000]$$

$$\max(x) = [1000, 1110, 820, 10000, 0.15, 500, 2000, 12000]$$

The above equations indicate that all low-fidelity functions have nonlinear model form discrepancy. To roughly quantify these discrepancies, we follow the same procedure as in the previous example and calculate RRMSEs, see **Table 3**. As it can be seen, the accuracy of the models increases with $i$.

**Table 3 Relative accuracy of functions for borehole problem:** The functions are listed in increasing order with respect to accuracy, with $y_{l_3}(x)$ being the most accurate by a significant margin.

|  | $y_{l_1}(x)$ | $y_{l_2}(x)$ | $y_{l_3}(x)$ |
|---|---|---|---|
| *RRMSE* | 3.6671 | 1.3688 | 0.36232 |

We consider various amounts of available low-fidelity data, with and without noise. We also use a few combinations for training LMGP based on the selected datasets or how data sources are encoded. The results are summarized in **Figure 12** where, once again, LMGP convincingly outperforms GP in high-fidelity emulation, especially with noisy data, see **Figure 12 (b)** and **(d)**. The overall trends in performance between strategies for LMGP are consistent across the various cases, with LMGP fit to only one low-fidelity source performing worse than LMGP fit to all data sources and with LMGP$_{s\ All}$ specifically performing the best. LMGP$_{m\ All}$ yields inconsistent results with $n_l = 50$ or $n_l = 100$, especially in the latter case where the box plots have stretched to include the outliers. This behavior is due to overfitting and the fact that there are many latent positions that must be placed in the latent space via a simple matrix-based map (256 positions and 32 elements in the $A$ matrix). Note that even with these inconsistencies, LMGP$_{m\ All}$ performs better that GP, LMGP$_{s\,l_1}$, and LMGP$_{m\,l_1}$ which indicates that using more than two datasets in fusion is indeed beneficial.

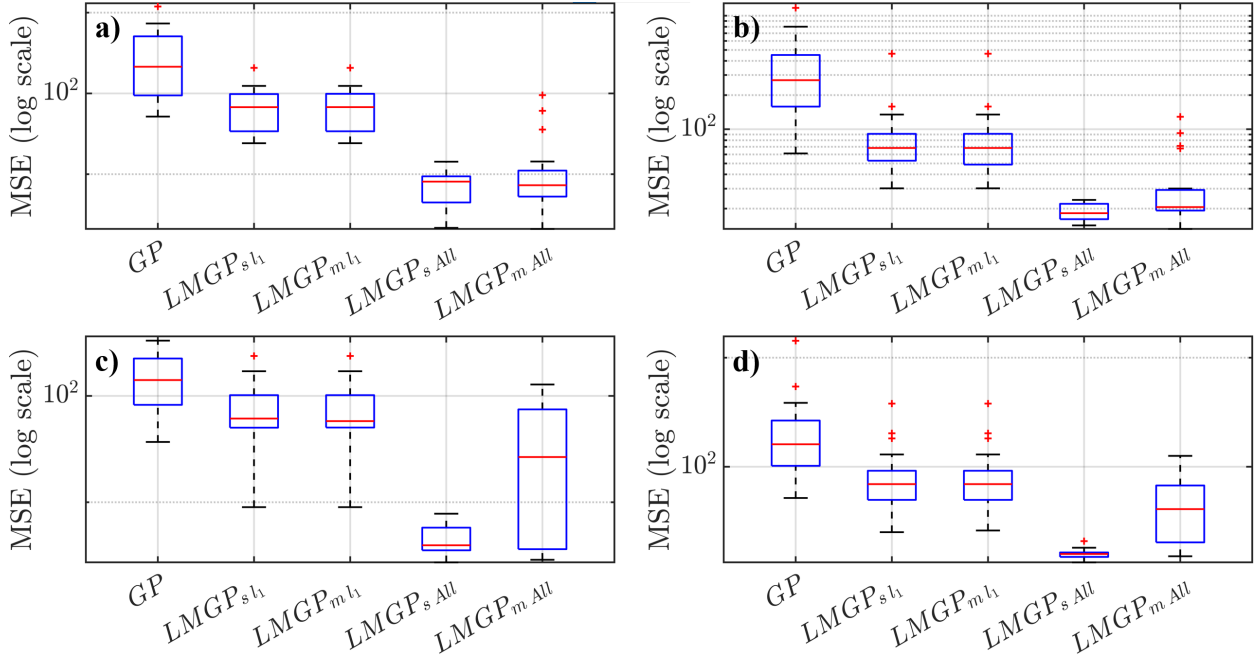

**Figure 12** High-fidelity emulation performance for the borehole problem: **(a)** $n_h = 15$, $n_{l_1} = n_{l_2} = n_{l_3} = 50$, $\sigma^2 = 0$: LMGP strategies that use all data sources perform better than those using only one data source, with LMGP$_{s\,All}$ performing the best. **(b)** $n_h = 15$, $n_{l_1} = n_{l_2} = n_{l_3} = 50$, $\sigma^2 = 6.25$: LMGP$_{s\,All}$ performs noticeably better than other LMGP strategies for this case. **(c)** $n_h = 15$, $n_{l_1} = n_{l_2} = n_{l_3} = 100$, $\sigma^2 = 0$: LMGP$_{s\,All}$ again performs noticeably better than other LMGP strategies for this case. LMGP$_{m\,All}$ displays inconsistency in its estimates. **(d)** $n_h = 15$, $n_{l_1} = n_{l_2} = n_{l_3} = 100$, $\sigma^2 = 6.25$: LMGP$_{s\,All}$ again performs noticeably better than other LMGP strategies for this case. LMGP$_{m\,All}$ again displays inconsistency in its estimates.

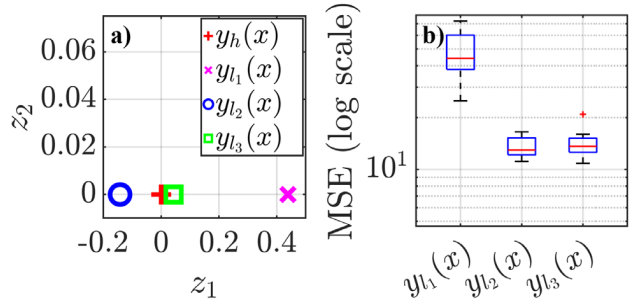

**Figure 13** Effects of correlations between data sources for borehole example: **(a) Latent space:** The latent space for one sample repetition of LMGP fit to all available data for the borehole function with $n_h = 15$, $n_{l_1} = n_{l_2} = n_{l_3} = 50$, $\sigma^2 = 6.25$. While the individual latent spaces are different for each repetition, the relative latent distances are consistent both for different repetitions and for different amounts of data/noise. **(b) Low-fidelity MSEs:** Low fidelity prediction accuracy is better for $y_{l_2}(x)$ and $y_{l_3}(x)$ than for $y_{l_1}(x)$.

The learned latent space for LMGP$_{s\,All}$ which is the most challenging version of this problem (noisy samples, fewest available data) is shown in **Figure 13 (a)** which clearly indicates that relative distances among the positions match with the relative accuracy between the low- and high-fidelity sources: The position for $y_{l_3}(x)$ is very close to that for $y_h(x)$, so LMGP weighs data from $y_{l_3}(x)$ heavily when emulating $y_h(x)$ and vice versa. The position for $y_{l_2}(x)$ is also close to both $y_h(x)$ and $y_{l_3}(x)$, but it is relatively more distant from $y_h(x)$ compared to $y_{l_3}(x)$.

Like in our first example, prediction performance on the low-fidelity sources for LMGP$_{s\ All}$, shown in **Figure 13 (b)**, follows a similar trend to data source accuracy, i.e., it is best for $y_{l_2}(x)$ and $y_{l_3}(x)$ and worst for $y_{l_1}(x)$ which is the least accurate source. As we mentioned before, we expect prediction accuracy to be high on sources whose latent positions are close together or form a cluster. In this case, $y_h(x)$, $y_{l_2}(x)$, and $y_{l_3}(x)$ form a cluster and as such we see that MSEs for $y_{l_2}(x)$ and $y_{l_3}(x)$ are much lower than those for $y_{l_1}(x)$.

## 4.2 Calibration Results

We compare our calibration approach to that of KOH detailed in section 2.2, by considering three test cases with varying degrees of complexity. Note that, while LMGP can simultaneously assimilate and calibrate any number of sources, KOH's approach only works with two datasets at a time and relies on repeating the process for as many times as there are low-fidelity sources.

For our first calibration problem, we consider data drawn from simple one-dimensional analytical functions:

$$y_h(x) = \frac{1}{0.1x^3+x^2+x+10}, \quad -2 \leq x \leq 3 \quad\quad \text{Eq. 23.1}$$

$$y_{l_1}(x) = \frac{1}{0.1x^3+\theta x^2+1.5x+10.5}, \quad -2 \leq x \leq 3 \text{ and } -1 \leq \theta \leq 2 \quad\quad \text{Eq. 23.2}$$

$$y_{l_2}(x) = \frac{1}{\theta x^2+x+10}, \quad -2 \leq x \leq 3 \text{ and } -1 \leq \theta \leq 2 \quad\quad \text{Eq. 23.3}$$

where we consider $\theta^* = 1$. Note that both low-fidelity sources have model form error, with $y_{l_2}(x)$ being more accurate than $y_{l_1}(x)$ over the input range when $\theta = \theta^*$ despite omitting the $x^3$ term (see **Table 4**).

**Table 4 Relative accuracy of functions for simple calibration problem:** We find the RRMSE in calibration problems using the same method as before but with the calibration parameters fixed to their true values at all input points. Both low-fidelity functions are relatively accurate, with $y_{l_2}(x)$ more accurate than $y_{l_1}(x)$.

|  | $y_{l_1}(x)$ | $y_{l_2}(x)$ |
|---|---|---|
| RRMSE | 0.22241 | 0.1285 |

We show high-fidelity emulation performance for this problem in **Figure 14** where, similar to Sec. 4.1, LMGPs are trained under various settings in terms of which data sources are selected and how they are encoded. As it can be observed LMGP performs on par with or better than KOH's approach in high-fidelity emulation accuracy for all cases, and LMGP$_{s\ All}$ offers the most consistent performance for most cases. LMGP also performs particularly well in the cases with noise, see **Figure 14 (b)** and **(d)**. Despite the inaccuracy of $y_{l_2}(x)$, LMGP fit to all data sources offers the most accurate emulation in all cases.

We next show calibration performance in **Figure 15** where LMGP$_{s\ All}$ consistently outperforms KOH in both accuracy and consistency, especially in the noiseless cases, see **Figure 15 (a)** and **(c)**. Notably, KOH's approach fit with $y_{l_2}(x)$ yields biased estimates. With noise and little data, see **Figure 15 (b)**, neither LMGP nor KOH's approach are able to obtain a very consistent estimate for the calibration parameter across the repetitions. When more low-fidelity data are provided, see

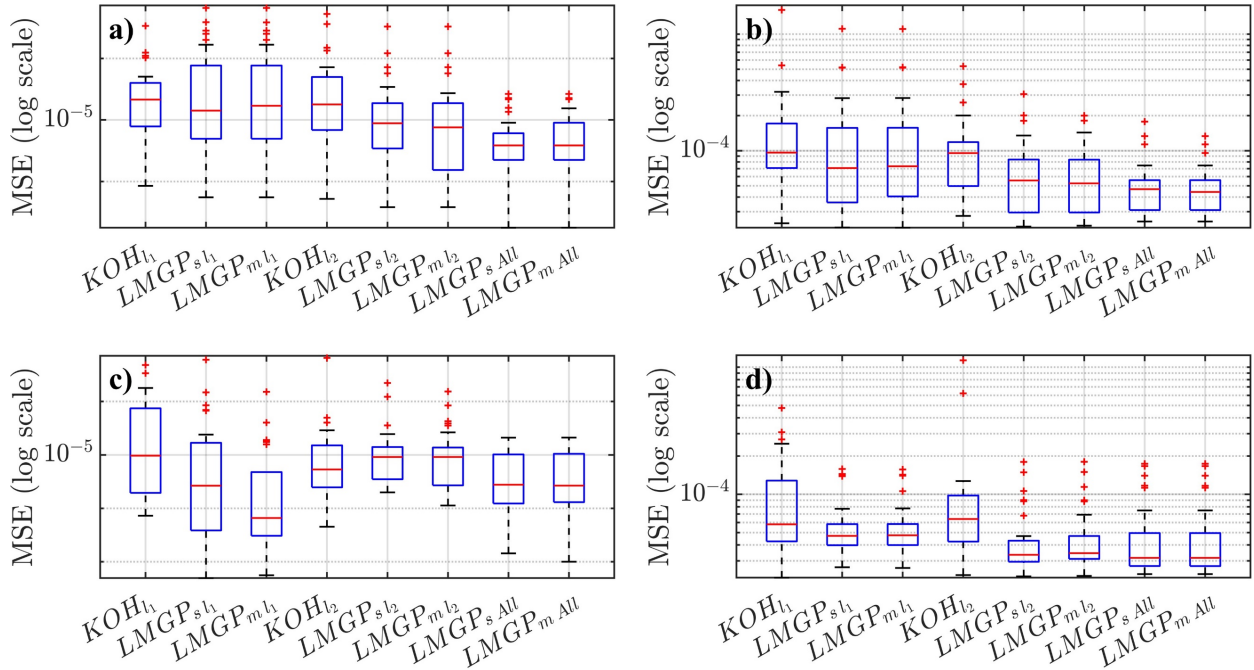

**Figure 14 High-fidelity emulation performance: (a)** $n_h = 3$, $n_{l_1} = n_{l_2} = n_{l_3} = 15$, $\sigma^2 = 0$: LMGP strategies generally perform better than KOH's approach, with LMGP$_{s\ All}$ performing the best. Estimates for all strategies except LMGP$_{s\ All}$ are fairly inconsistent. **(b)** $n_h = 3$, $n_{l_1} = n_{l_2} = n_{l_3} = 15$, $\sigma^2 = 2 \cdot 10^{-5}$: LMGP$_{s\ All}$ performs noticeably better than other LMGP strategies for this case (and better than KOH's approach). **(c)** $n_h = 3$, $n_{l_1} = n_{l_2} = n_{l_3} = 50$, $\sigma^2 = 0$: With the addition of more low-fidelity data, all approaches perform better. LMGP$_{s\ All}$ performs best by a very slim margin, and is more consistent in its performance than comparable strategies. **(d)** $n_h = 3$, $n_{l_1} = n_{l_2} = n_{l_3} = 50$, $\sigma^2 = 2 \cdot 10^{-5}$: With noise, LMGP$_{s\ l_2}$ performs nearly on par with LMGP$_{s\ All}$ and produces more consistent performance.

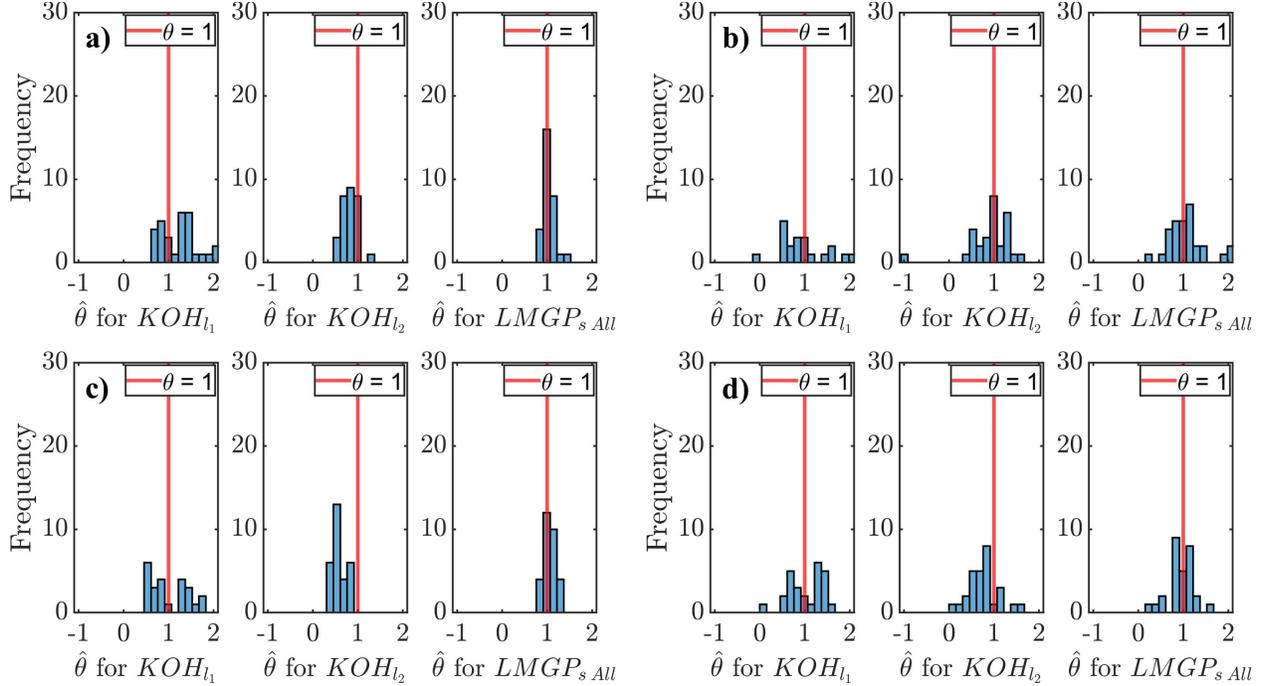

**Figure 15 Calibration performance: (a)** $n_h = 3$, $n_{l_1} = n_{l_2} = n_{l_3} = 15$, $\sigma^2 = 0$: LMGP offers consistent and unbiased estimates. KOH's approach suffers from bias and inconsistency. **(b)** $n_h = 3, n_{l_1} = n_{l_2} = n_{l_3} = 15, \sigma^2 = 2 \cdot 10^{-5}$: All approaches yield inconsistent estimates. **(c)** $n_h = 3, n_{l_1} = n_{l_2} = n_{l_3} = 50, \sigma^2 = 0$: Both KOH's approach and LMGP yield consistent estimates, but KOH's approach still suffers from bias. **(d)** $n_h = 3, n_{l_1} = n_{l_2} = n_{l_3} = 50, \sigma^2 = 2 \cdot 10^{-5}$: LMGP achieves higher consistency that KOH's approach with the addition of more low-fidelity data. LMGP's estimate is unbiased, while KOH's approach still yields biased estimates.

**Figure 15 (d)**, LMGP is able to leverage the additional low-fidelity data to find a consistent estimate for $\theta$ while KOH's approach does not improve in consistency.

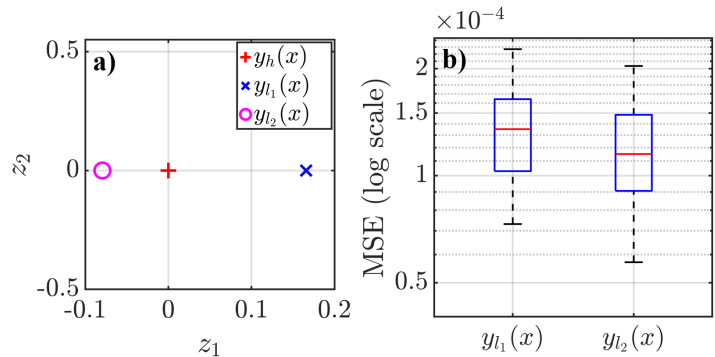

**Figure 16 Effects of correlations between data sources: (a) Latent space:** The latent space for one sample repetition of LMGP fit to all available data with $n_h = 3, n_{l_1} = n_{l_2} = n_{l_3} = 15, \sigma^2 = 2 \times 10^{-5}$. While the individual latent spaces are different for each repetition, the relative latent distances are consistent both for different repetitions and for different amounts of data/noise. **(b) Low-fidelity MSEs:** Low fidelity prediction accuracy is better for $y_{l_2}(x)$ than for $y_{l_1}(x)$.

We show the latent space from fitting LMGP to the most challenging version of this problem, i.e., $n_h = 3$, $n_{l_1} = n_{l_2} = 15$, $\sigma^2 = 2 \times 10^{-5}$. As demonstrated in **Figure 16 (a)** LMGP is able to accurately infer the correlations with only 3 noisy high-fidelity samples as the relative latent distances match the relative accuracies of the data sources. Thus, we expect the low-fidelity performance to be better for $y_{l_2}(x)$ than for $y_{l_1}(x)$ as the position for $y_{l_2}(x)$ is relatively closer to $y_h(x)$ which means that LMGP leverages more information from $y_h(x)$ in predicting $y_{l_2}(x)$ than in predicting $y_{l_1}(x)$. We assess the veracity of our expectation by examining low-fidelity prediction performance in **Figure 16 (b)** which indicates that prediction performance is indeed better for $y_{l_2}(x)$ than for $y_{l_1}(x)$.

Next, we reconsider the example in Eq. 20 where $\theta^* = \pi$ and $\theta^* = 10\pi$ are the two valid choices for the true calibration parameter as discussed in Sec. 3.2. We fit LMGP with two approaches to categorical variable selection and consider various amounts of available low-fidelity data all with noise (the noiseless case is considered in section 3.2).

The high-fidelity emulation performance is summarized in **Figure 17** which indicates that LMGP outperforms KOH's approach by a similar margin for each case. Notably, LMGP's performance is robust to the choice of categorical variable assignment for this problem as we see a similar variation in performance over repetitions between $\text{LMGP}_{s\ All}$ and $\text{LMGP}_{m\ All}$. We explain this by noting that since there are only two data sources, $\text{LMGP}_{m\ All}$ finds a total of $2^2 = 4$ latent positions with $(2 + 2) \times 2 = 8$ elements in $A$ which indicates that overfitting should not be a concern.

The estimates of the calibration parameters are provided in **Figure 18** and indicate that the estimation consistency in both approaches increases as $n_l$ is increased from 30 to 200. This increase is more prominent for LMGP. However, while LMGP converges on $\theta = 10\pi$, KOH's approach's estimates are approximately evenly split between $\pi$ and $10\pi$. This behavior is because the $L2$ distance of $\sin(10\pi x)$ and $\sin(\pi x)$ from $y_h(x)$ is the same and hence KOH's approach cannot favor one over the other [21, 50, 51]. As explained in Sec. 3.2, in this case LMGP converges at $\theta = 10\pi$ as this choice provides not only a simpler discrepancy but also enables learning the high frequency nature of $y_h(x)$.

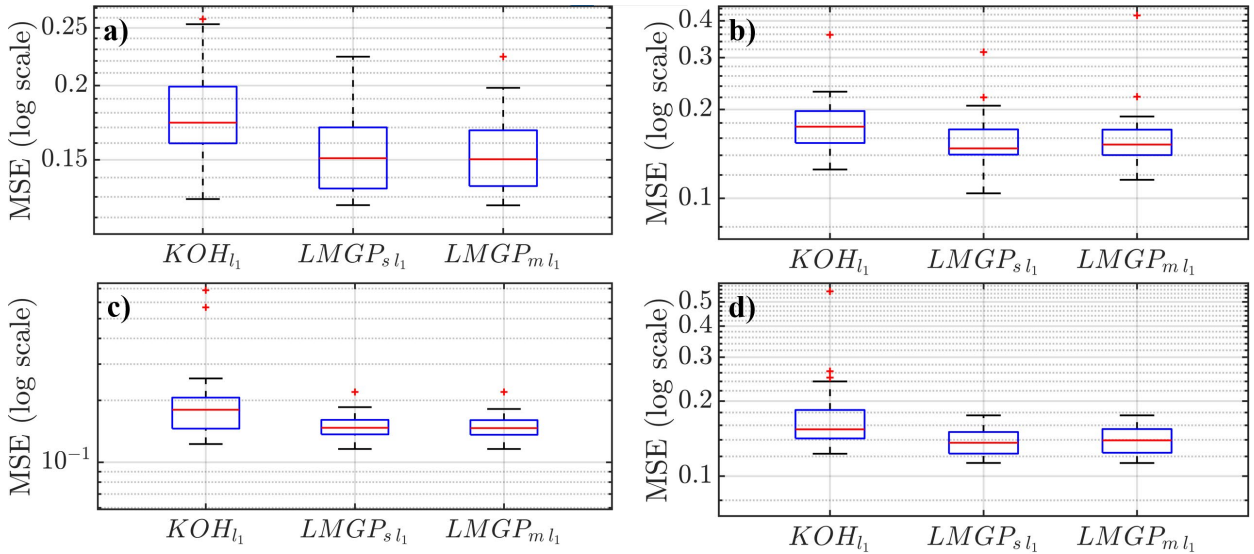

**Figure 17** High-fidelity emulation performance for sin wave example: **(a)** $n_h = 30, n_l = 30, \sigma^2 = .09$, **(b)** $n_h = 30, n_l = 60, \sigma^2 = .09$, **(c)** $n_h = 30, n_l = 100, \sigma^2 = .09$, and **(d)** $n_h = 30, n_l = 200, \sigma^2 = .09$. LMGP outperforms KOH's approach by a similar margin in all cases.

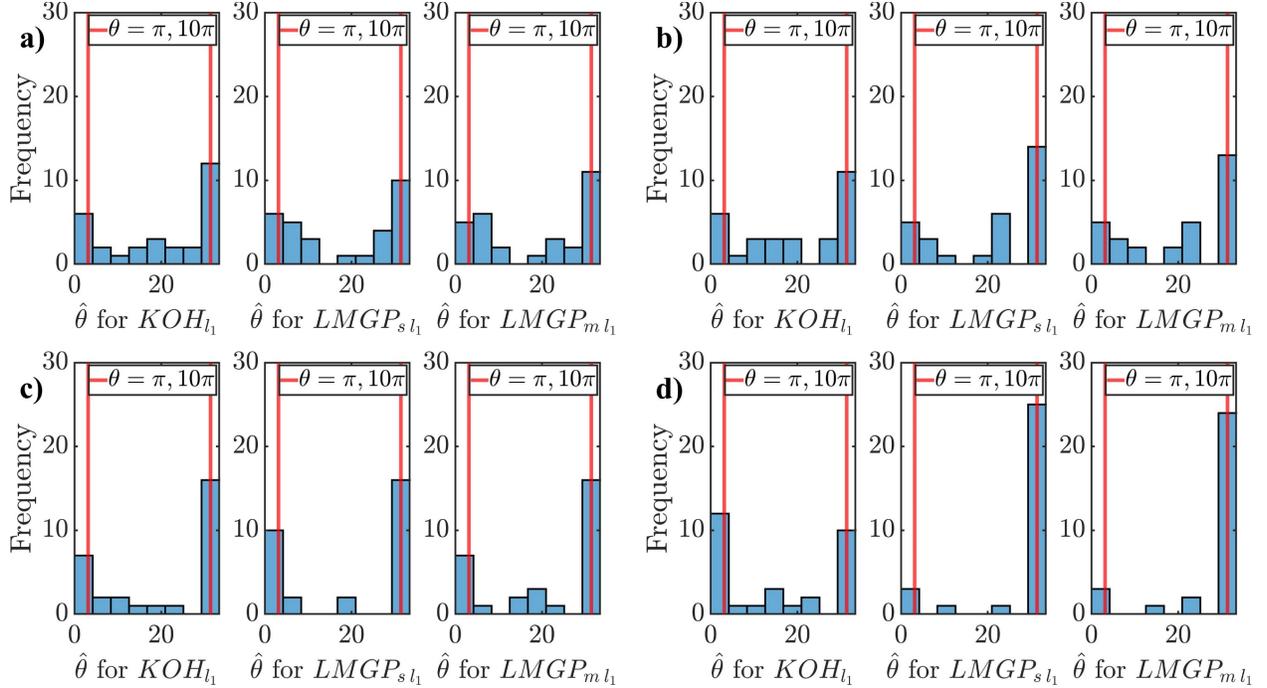

**Figure 18 Calibration performance for sin wave problem:** (a) $n_h = 30$, $n_l = 30$, $\sigma^2 = .09$, (b) $n_h = 30$, $n_l = 60$, $\sigma^2 = .09$, (c) $n_h = 30$, $n_l = 100$, $\sigma^2 = .09$, and (d) $n_h = 30$, $n_l = 200$, $\sigma^2 = .09$.

Finally, we show histograms of latent distances learned by LMGP in **Figure 19**. The trends are quite similar to those seen in section 3.2, with the latent distances being close to 0 for low amounts of low-fidelity data and converging on 0.5 as the amount of data is increased. When high-fidelity data is insufficient to learn the high-frequency behavior of $y_h(x)$, LMGP treats the high-frequency behavior as noise and finds $y_h(x) \approx \sin(\pi x)$. When low-fidelity data are also insufficient, LMGP cannot learn the behavior of $y_l(x)$ at high frequencies (i.e., for large $\theta$). Thus, LMGP finds $\theta = \pi$ which implies $y_l(x) = \sin(\pi x)$, i.e., no model form error and a corresponding latent distance near zero. With sufficient low-fidelity data, however, LMGP learns the behavior of $y_l(x)$ for large $\theta$ and finds that $\theta = 10\pi$ yields a less complex discrepancy between $y_h(x)$ and $y_l(x)$.

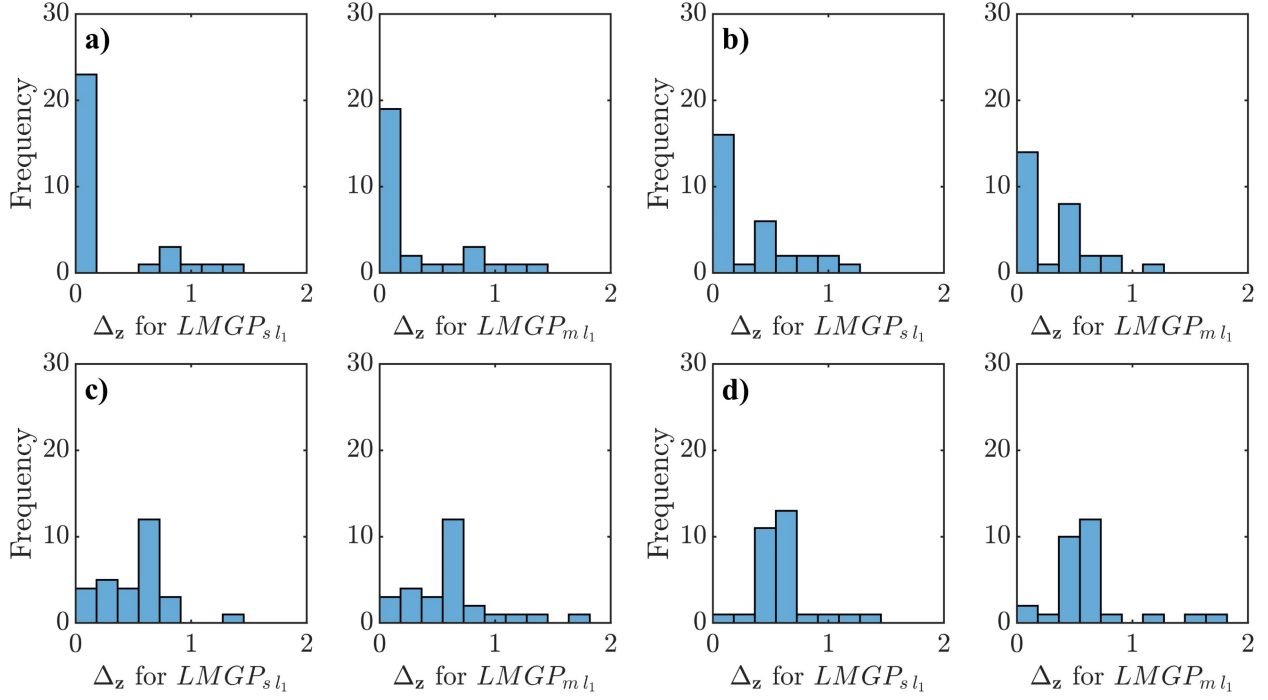

**Figure 19 Histogram of latent distances for sin wave problem: (a)** $n_h = 30$, $n_l = 30$, $\sigma^2 = .09$, **(b)** $n_h = 30$, $n_l = 60$, $\sigma^2 = .09$, **(c)** $n_h = 30$, $n_l = 100$, $\sigma^2 = .09$, **(d)** $n_h = 30$, $n_l = 200$, $\sigma^2 = .09$.

We now revisit the borehole problem from section 4.1, this time adapted as a calibration problem to explore the effects of both nonlinear model form error and high-dimensional inputs. We begin with data drawn from the following functions:

$$y_h(x) = \frac{2\pi T_u (H_u - H_l)}{ln\left(\frac{r}{r_w}\right)\left(1 + \frac{2 \cdot 1500 \cdot T_u}{ln(r/r_w) r_w^2 K_w} + \frac{T_u}{250}\right)} \qquad \text{Eq. 24.1}$$

$$y_{l1}(x) = \frac{2\pi \times 500 \times (0.993 \times H_u - H_l)}{0.95 \times ln\left(\frac{r}{r_w}\right)\left(1 + \frac{2\theta_2 \times 500}{ln(r/r_w) r_w^2 K_w} + \frac{500}{\theta_1}\right)} \qquad \text{Eq. 24.2}$$

$$y_{l2}(x) = \frac{2\pi \times 500 \times (H_u - 1.045 \times H_l)}{ln\left(\frac{r}{r_w}\right)\left(1 + \frac{2\theta_2 \times 500}{ln(r/r_w) r_w^2 K_w} + \frac{500}{\theta_1}\right)} \qquad \text{Eq. 24.3}$$

$$x^T = [T_u, H_u, H_l, r, r_w, K_w], \quad \boldsymbol{\theta}^T = [\theta_1, \theta_2],$$

$$\min(x) = [100, 990, 700, 100, 0.05, 6000],$$

$$\max(x) = [1000, 1110, 820, 10000, 0.15, 12000],$$

$$\min(\boldsymbol{\theta}^T) = [10, 1000], \max(\boldsymbol{\theta}^T) = [500, 2000]$$

where we consider $\boldsymbol{\theta}^{T*} = [250, 1500]$. Note that both low-fidelity sources have model form error with $y_{l_1}(x)$ being more accurate than $y_{l_2}(x)$ over the input range when the true calibration parameters are used, see **Table 5**, and that the input $T_u$ has been omitted and replaced by a constant in both low-fidelity functions.

**Table 5 Relative accuracy of functions for borehole calibration problem:** Both low-fidelity functions are relatively accurate, with $y_{l_2}(x)$ less accurate than $y_{l_1}(x)$.

|  | $y_{l_1}(x)$ | $y_{l_2}(x)$ |
|---|---|---|
| RRMSE | 0.049219 | 0.19838 |

We hold $n_h = 25$ and $n_l = 100$ constant and examine two cases, one without noise and one with noise applied to samples ($\sigma^2 = 100$ with $\text{Range}(y_h(x)) \approx 974$ over the input range) and again fit LMGP with various strategies. In both cases, LMGP convincingly outperforms KOH's approach in high-fidelity emulation, see **Figure 20**. Notably, LMGP outperforms KOH's approach given equivalent access to data, e.g., $\text{LMGP}_{s\,l_1}$ versus $\text{KOH}_{l_1}$. LMGP's performance is also robust to modeling choice, which we explain by noting that with three data sources the $t_m$ strategy for categorical variable selection yields $3^3 = 27$ latent positions and $2 \times (3 \times 3) = 18$ elements of $A$, i.e., the number of latent positions is on the same order of magnitude as the number of hyperparameters in $A$ and the size of the dataset is large relative to the number of hyperparameters.

As shown in **Figure 22 (a)** for the noiseless case, the latent positions found by $\text{LMGP}_{s\,All}$ show no model form error for $y_{l_1}(x)$ and little model form error for $y_{l_2}(x)$, i.e., LMGP mistakes model form error in $y_{l_1}(x)$ for noise since the error is so low. While these latent positions are not fully accurate as $y_{l_1}(x)$ does still have model form error, the relative distances to the data sources do correctly indicate which is more accurate. With noise, the relative distances to $y_h(x)$ are nearly the same for both low-fidelity sources, although $y_{l_1}(x)$ is slightly closer to $y_h(x)$ than $y_{l_2}(x)$, which indicates that LMGP has more difficulty determining the magnitudes of the errors in the low-fidelity data sources in this case. The magnitudes of the latent distances are quite small in both cases, which reflect the fact that both low-fidelity data sources are relatively accurate when calibrated appropriately.

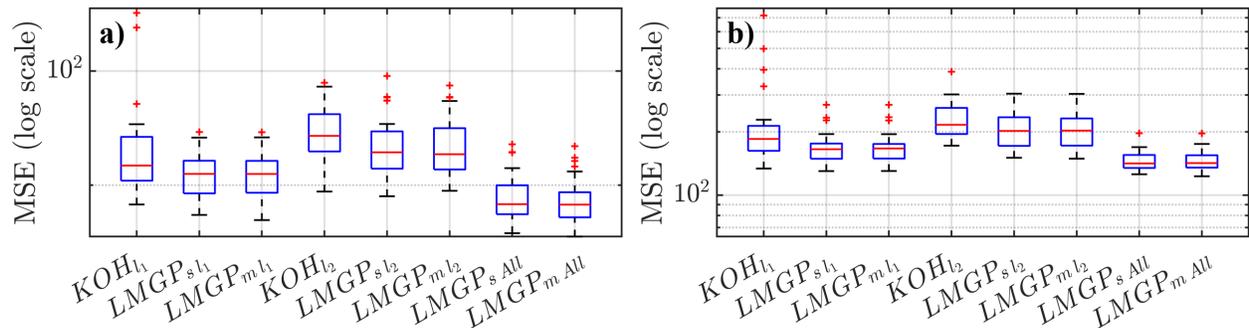

**Figure 20 High-fidelity emulation performance:** (a) $n_h = 25$, $n_{l_1} = n_{l_2} = n_{l_3} = 100$, $\sigma^2 = 0$: $\text{LMGP}_{m\,All}$ arguably performs better than $\text{LMGP}_{s\,All}$. (b) $n_h = 25$, $n_{l_1} = n_{l_2} = n_{l_3} = 100$, $\sigma^2 = 100$: Results with noise are quite similar to those without.

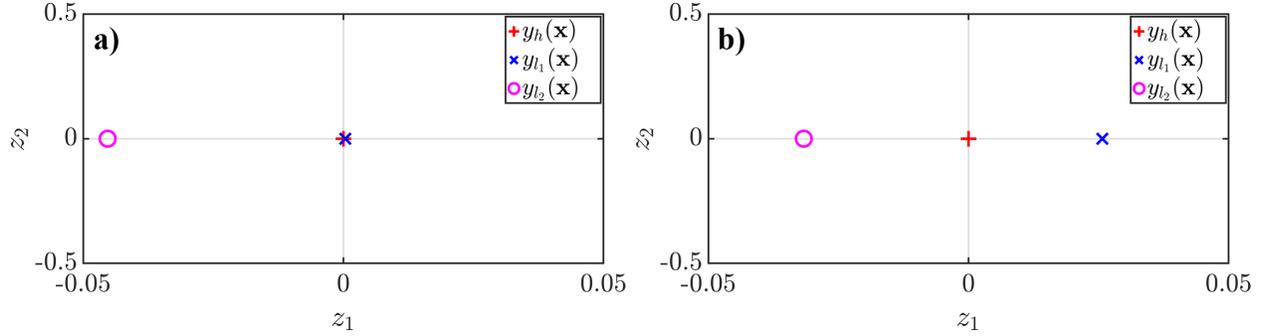

**Figure 22 Latent positions: (a)** $n_h = 25$, $n_{l_1} = n_{l_2} = n_{l_3} = 100$, $\sigma^2 = 0$: LMGP finds no model form error for $y_{l_2}(x)$ and instead mistakes it for noise. **(b)** $n_h = 25$, $n_{l_1} = n_{l_2} = n_{l_3} = 100$, $\sigma^2 = 100$: LMGP correctly finds little error for both sources, but is unable to accurately determine the relative magnitudes of those errors.

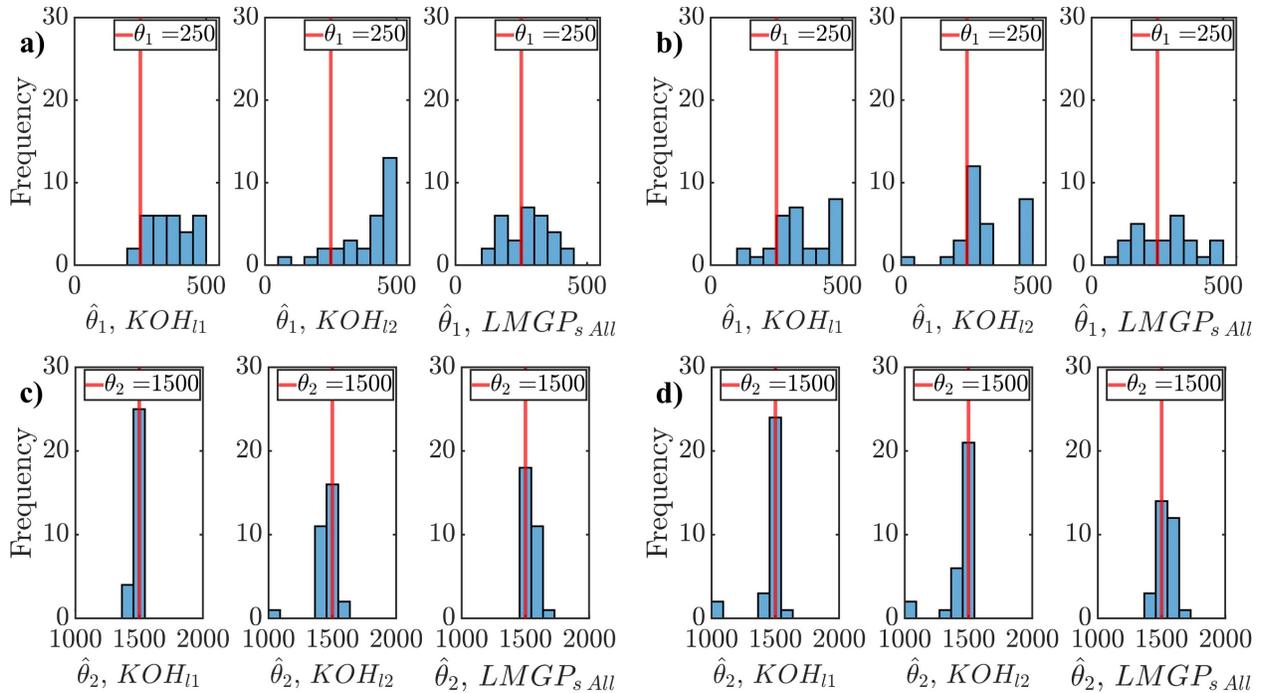

**Figure 21 Calibration performance: (a)** $\hat{\theta}_1$ for $n_h = 25$, $n_{l_1} = n_{l_2} = n_{l_3} = 100$, $\sigma^2 = 0$: KOH's approach produces biased estimates, while LMGP's estimates are centered on the correct parameter with high variance. **(b)** $\hat{\theta}_1$ for $n_h = 25$, $n_{l_1} = n_{l_2} = n_{l_3} = 100$, $\sigma^2 = 100$: KOH's approach again produces biased estimates, with the caveat that $KOH_{l_2}$ finds the correct parameter nearly half the time. **(c)** $\hat{\theta}_2$ for $n_h = 25$, $n_{l_1} = n_{l_2} = n_{l_3} = 100$, $\sigma^2 = 0$: All methods find the correct parameter consistently. $KOH_{l_1}$ finds the most accurate and consistent estimates, while $KOH_{l_2}$ has some outliers. **(d)** $\hat{\theta}_2$ for $n_h = 25$, $n_{l_1} = n_{l_2} = n_{l_3} = 100$, $\sigma^2 = 100$: Both of KOH's approaches have outliers, but estimate the correct parameter more consistently than LMGP.

Calibration performance, shown in **Figure 21**, reveals inconsistent performance in estimating $\theta_1$ but consistent estimates for $\theta_2$ for both LMGP and KOH's approach in all three cases. We explain this by noting that the main sensitivity indices (calculated using 10,000 inputs sampled via sobol sequence) for $\theta_1$ and $\theta_2$ are on the order of $10^{-4}$ and $10^{-1}$ respectively for the low-fidelity functions, i.e., variation in $\theta_1$ has very little effect on their outputs. Therefore, we expect $\theta_1$ to be very difficult to estimate. While LMGP's estimates for $\theta_1$ suffer from high variance, the distributions are centered on the true parameter for both cases. By contrast, KOH's approach produces biased estimates in all cases, although $KOH_{l_2}$ guesses nearly the correct parameter almost half the time in the case with noise, see **Figure 21 (b)**. Both methods estimate $\theta_2$ quite accurately and consistently. KOH's approach has lower variance in its estimates but more outliers when using $y_{l_2}(x)$ compared to LMGP's estimates using all data sources.

## 5  Conclusion

In this paper, we present a novel latent-space based approach for data fusion (i.e., multi-fidelity modeling and calibration) via latent map Gaussian processes or LMGPs. Our approach offers unique advantages that can benefit engineering design in a number of ways such as improved accuracy and consistency compared to competing methods for data fusion. Additionally, LMGP learns a latent space where data sources are embedded with points whose distances can shed light on not only the relations among data sources, but also potential model form discrepancies. These insights can guide diagnostics or determine which data sources cannot be trusted.

Implementation and use of our data fusion approach is quite straightforward as it primarily relies on modifying the correlation function of traditional GPs and assigning appropriate priors to the datasets. LMGP-based data fusion is also quite flexible in terms of the number of data sources. In particular, since we can assimilate multiple datasets simultaneously, we improve prediction performance and decrease non-identifiability issues that typically arise in calibration problems.

Since LMGPs are extensions of GPs, they are not directly applicable to extrapolation or big/high-dimensional data. However, extensions of GPs that address these limitations (see [27, 37, 44-47, 52] for some examples) can be incorporated into LMGPs. In our examples, we assumed all data sources are noisy and hence used a single parameter to estimate the noise. To consider different (unknown) noise levels, we need to have a parameter for each data source. We also note that the performance of LMGP in fusing small data can be greatly improved by endowing its parameters with priors and using Bayes' rule for inference. In this case, the latent space will have a probabilistic nature, the trained model will be more robust to overfitting, and prediction uncertainties will be more accurate. These and other directions will be investigated in our future works.

Lastly, we note that the proposed method can be directly applied to multi-response datasets with no modifications. To apply LMGP, we would treat each response as a separate dataset and apply the multi-fidelity method we present directly. However, with this strategy each 'data source' would have the exact same set of input points, which will most likely cause numerical issues. While LMGP can be applied to multi-response datasets with some modifications (which may be presented in a future paper), the user should bear in mind that we do not necessarily *a priori* expect any level of correlation between the responses whereas with multi-fidelity problems we expect (but do not necessarily have) some correlation as all sources model the same system. Thus, we would recommend fitting LMGP to all responses and examining the latent space to see which responses are well-correlated. Then, fit individual emulators to uncorrelated responses while fitting an LMGP to whichever groups of responses that are correlated with each other.

# 6 Acknowledgements

This work was supported by the Early Career Faculty grant from NASA's Space Technology Research Grants Program (award number 80NSSC21K1809).